\documentclass{article}

\usepackage{natbib}
\usepackage{frExamplee}
\usepackage{graphicx}
\usepackage{setspace}

\usepackage{url}
\usepackage{textcomp}
\usepackage{gensymb}
\usepackage{blindtext}
\usepackage{amsmath}
\usepackage{amssymb}
\usepackage{algorithm}
\usepackage[usenames,dvipsnames,svgnames,table]{xcolor}
\usepackage[noend]{algpseudocode}
\usepackage[acronym]{glossaries} 
\usepackage[caption=false]{subfig}
\usepackage[all]{nowidow}
\usepackage{etoolbox}
\usepackage{threeparttable}
\usepackage{booktabs}
\usepackage{multirow}
\usepackage{adjustbox}
\usepackage{cellspace}
\usepackage{siunitx}
\usepackage{titlesec}

\titlespacing*{\section}{0pt}{0.1\baselineskip}{0.1\baselineskip}
\titlespacing*{\subsection}{0pt}{0.1\baselineskip}{0.1\baselineskip}
\titlespacing*{\subsubsection}{0pt}{0.1\baselineskip}{0.1\baselineskip}

\makeatletter
\AfterEndEnvironment{algorithm}{\let\@algcomment\relax}
\AtEndEnvironment{algorithm}{\kern2pt\hrule\relax\vskip3pt\@algcomment}
\let\@algcomment\relax
\newcommand\algcomment[1]{\def\@algcomment{\footnotesize#1}}

\renewcommand\fs@ruled{\def\@fs@cfont{\bfseries}\let\@fs@capt\floatc@ruled
	\def\@fs@pre{\hrule height.8pt depth0pt \kern2pt}%
	\def\@fs@post{}%
	\def\@fs@mid{\kern2pt\hrule\kern2pt}%
	\let\@fs@iftopcapt\iftrue}
\makeatother

\newcommand\psecref[1]{\mbox{(Section \ref{sec:#1})}}
\newcommand\tsecref[1]{\mbox{Section \ref{sec:#1}}}
\newcommand\figref[1]{\mbox{(Fig. \ref{fig:#1})}}
\newcommand\tfigref[1]{\mbox{Fig. \ref{fig:#1}}}

\newcommand\ptblref[1]{\mbox{(Table \ref{tbl:#1})}}
\newcommand\ttblref[1]{\mbox{Table \ref{tbl:#1}}}

\newcommand\seclabel[1]{\label{sec:#1}}
\newcommand\figlabel[1]{\label{fig:#1}}

\newcommand\tbllabel[1]{\label{tbl:#1}}

\usepackage[utf8]{inputenc}

\title{\emph{Osprey:} Multi-Session Autonomous Aerial Mapping with LiDAR-based SLAM and Next Best View Planning}

\author{
Rowan Border \\
Oxford Robotics Institute (ORI) \\
Department of Engineering Science \\
University Of Oxford \\
\texttt{rborder.robots@gmail.com} \\
\And
Nived Chebrolu \\
Oxford Robotics Institute (ORI) \\
Department of Engineering Science \\
University of Oxford \\
\texttt{nived@robots.ox.ac.uk} \\
\And
Yifu Tao \\
Oxford Robotics Institute (ORI) \\
Department of Engineering Science \\
University of Oxford \\
\texttt{yifu@robots.ox.ac.uk} \\
\AND
Jonathan D. Gammell \\
Oxford Robotics Institute (ORI) \\
Department of Engineering Science \\
University of Oxford \\
\texttt{gammell@robots.ox.ac.uk} \\
\And
Maurice Fallon \\
Oxford Robotics Institute (ORI) \\
Department of Engineering Science \\
University of Oxford \\
\texttt{mfallon@robots.ox.ac.uk} \\
}

\begin{document}

\maketitle

\newacronym{inf}{IG}{Information Gain}
\newacronym{nbv}{NBV}{Next Best View}
\newacronym{see}{SEE}{Surface Edge Explorer}
\newacronym{prm}{PRM}{Probabilistic Road Map}
\newacronym{cnn}{CNN}{Convolutional Neural Network}
\newacronym{gng}{GNG}{Growing Neural Gas}
\newacronym{pca}{PCA}{Principal Component Analysis}
\newacronym{pga}{PGA}{Principal Geodesic Analysis}
\newacronym{pns}{PNS}{Principal Nested Spheres}
\newacronym{pngs}{PNGS}{Principal Nested Great Spheres}
\newacronym{hpr}{HPR}{Hidden Point Removal}
\newacronym{slsqp}{SLSQP}{Sequential Least-Squares Quadratic Programming}
\newacronym{gp}{GP}{Gaussian Process}
\newacronym{ros}{ROS}{Robot Operating System}
\newacronym{dbscan}{DBSCAN}{Density-Based Spatial Clustering of Applications with Noise}
\newacronym{icp}{ICP}{Iterative Closest Point}
\newacronym{oumnh}{OUMNH}{Oxford University Museum of Natural History}
\newacronym{tsp}{TSP}{Travelling Salesman Problem}
\newacronym{ugv}{UGV}{Unmanned Ground Vehicle}
\newacronym{uav}{UAV}{Unmanned Aerial Vehicle}
\newacronym{auv}{AUV}{Autonomous Underwater Vehicle}
\newacronym{rrt}{RRT}{Rapidly-exploring Random Tree}
\newacronym{rrg}{RRG}{Rapidly-exploring Random Graph}
\newacronym{rgg}{RGG}{Random Geometric Graph}
\newacronym{mcmc}{MCMC}{Markov Chain Monte Carlo}
\newacronym{tsdf}{TSDF}{Truncated Signed Distance Field}
\newacronym{ompl}{OMPL}{Open Motion Planning Library}
\newacronym{aitstar}{AIT*}{Adaptively Informed Trees}
\newacronym{knn}{\textit{k}-NN}{\textit{k}-Nearest Neighbours}
\newacronym{imu}{IMU}{Inertial Measurement Unit}
\newacronym{ori}{ORI}{Oxford Robotics Institute}
\newacronym{slam}{SLAM}{Simultaneous Localisation and Mapping}
\newacronym{vilens}{VILENS}{Visual Inertial Legged/Lidar Navigation System}
\newacronym{mvs}{MVS}{Multi View Stereo}
\newacronym{ekf}{EKF}{Extended Kalman Filter}
\newacronym{gmm}{GMM}{Gaussian Mixture Model}
\newacronym{nerf}{NeRF}{Neural Radiance Field}
\newacronym{tls}{TLS}{terrestrial laser scanner}

\begin{abstract}          
Aerial mapping systems are important for many surveying applications (e.g., industrial inspection or agricultural monitoring). Aerial platforms that can fly GPS-guided preplanned missions semi-autonomously are already widely available but fully autonomous systems can significantly improve efficiency. Autonomously mapping complex 3D structures requires a system that performs online mapping and mission planning. This paper presents \textit{Osprey}, an autonomous aerial mapping system with state-of-the-art multi-session LiDAR-based mapping capabilities. It enables a non-expert operator to specify a bounded target area that the aerial platform can then map autonomously over multiple flights.  Field experiments with Osprey demonstrate that this system can achieve greater map coverage of large industrial sites than manual surveys with a pilot-flown aerial platform or a \gls{tls}. Three sites, with a total ground coverage of $2528$ \SI{}{\meter\squared} and a maximum height of $27$~m, were mapped in separate missions using $112$ minutes of autonomous flight time. True colour maps were created from images captured by Osprey using pointcloud and NeRF reconstruction methods. These maps provide useful data for structural inspection tasks.  
\end{abstract}

\glsresetall

\begin{figure}[tp]
	\centering
	\captionsetup[subfigure]{labelformat=empty}
	\subfloat[]{\includegraphics[width=0.6\linewidth]{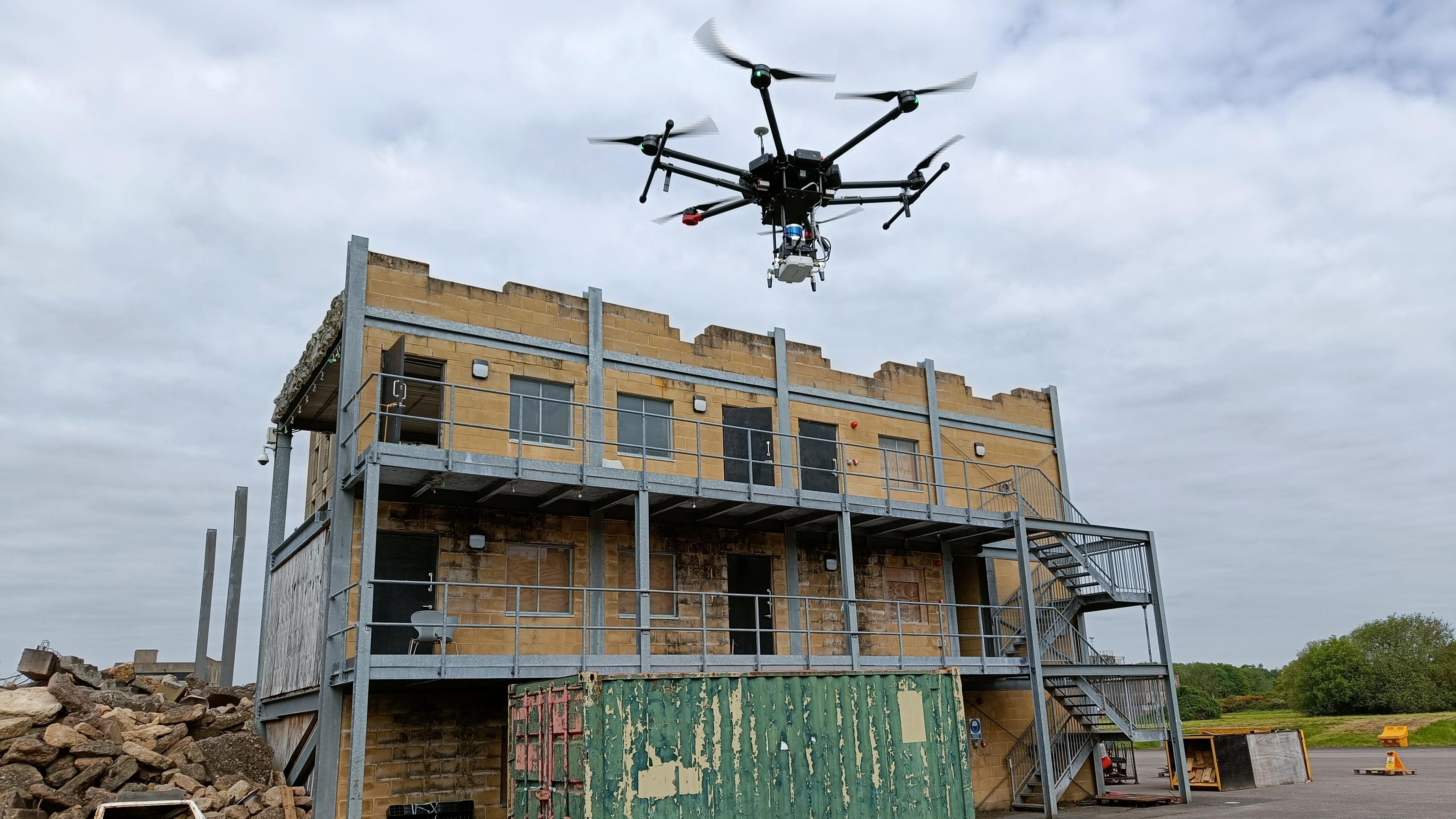}}
	\vspace{-4ex}
	\caption{The Osprey autonomous aerial mapping system surveying a multi-building industrial site.}
	\figlabel{marquee}
\end{figure}

\section{Introduction}

Autonomous aerial mapping systems can provide crucial information for a variety of applications, including industrial inspection, construction surveying and agricultural monitoring. For some tasks (e.g., terrain mapping) it is sufficient for an aerial platform with GPS navigation to fly along a fixed preplanned trajectory while capturing data with a camera or LiDAR. Many semi-autonomous aerial systems without online mapping or mission planning capabilities already provide this functionality (e.g., the DJI Mavic 3). Mapping structures (e.g., buildings or bridges) is more complex and requires a fully autonomous system capable of adaptively planning a flight trajectory online from live sensor measurements. This allows the aerial platform to safely manoeuvre around unknown environments and capture complete maps of complex structures.       

Fully autonomous aerial mapping systems can map large structures more quickly than ground-based systems (e.g., a \gls{tls} or robot platform) and obtain greater coverage by flying above and around tall structures that would otherwise be unobservable. Their autonomy enables them to operate without oversight or communication from a human pilot, thereby reducing the cost of data acquisition as well as allowing them to map hazardous environments that are not safely accessible (e.g., nuclear waste facilities). The ultimate aim for many of these systems is for an untrained user to specify an area or object of interest that the aerial platform can then autonomously map without human intervention.

The industrial importance of fully autonomous aerial mapping is already being demonstrated by commercial systems (e.g., Emesent Autonomy, Leica BLK2FLY and Skydio 3D Scan). These are being deployed to survey mining tunnels, tall buildings and cellphone towers. Mapping missions typically exceed battery flight time and as a result some commercial systems can incrementally build a map over multiple flights.

Research on fully autonomous aerial mapping systems has advanced significantly in the past decade. \citet{Shen2012} presented a 2D LiDAR-based autonomous aerial mapping system and demonstrated it in an indoor environment. \citet{Heng2014} presented a vision-based system and demonstrated it mapping small indoor and outdoor environments. \cite{Yoder2016} presented a system capable of mapping large outdoor structures with a 2D LiDAR and demonstrated it surveying a bridge. Autonomous mapping systems have since been developed using a variety of different platforms, sensor payloads and mapping pipelines. They have been demonstrated indoors \citep[e.g.,][]{Bircher2016,Bircher2018a}, on small outdoor structures  \citep[e.g.,][]{Teixeira2018}, large outdoor structures \citep[e.g.,][]{Song2020a,Song2021} and underground \citep[e.g.,][]{Papachristos2019}. The capabilities of state-of-the-art research systems is exemplified by the aerial platforms fielded in the DARPA Subterranean Challenge \citep{Chung2023}, which autonomously mapped underground environments as part of heterogeneous robot teams.

This paper presents \textit{Osprey}, an autonomous aerial lidar-based mapping system capable of surveying large outdoor structures over multiple flights \figref{marquee}. We incorporated a custom sensor payload, called \emph{Frontier}, onto a DJI M600 aerial platform to deploy our autonomous multi-session mapping pipeline. The state-of-the-art capabilities of Osprey are demonstrated by its ability to autonomously map several different sites, including a large ($52$x$29$x$27$m) disaster site in $42$ minutes over consecutive two flights. 

The key contributions of this work are the development of a fully autonomous aerial mapping system, which integrates existing state-of-the-art approaches for multi-sensor odometry, multi-session LiDAR-based mapping, online mission planning, and efficient motion planning. We also present the results of field experiments demonstrating its capabilities for mapping large outdoor structures. A discussion of these experiments identifies outstanding challenges for autonomous aerial mapping and future research directions. To the best of our knowledge the only other research systems with comparable capabilities are those deployed in the DARPA Subterranean Challenge, which were demonstrated in underground and indoor environments (e.g., \citealp[by Team Explorer;][]{Scherer2022} and \citealp[Team CSIRO Data61;][]{Hudson2022}).

Field experiments with Osprey were conducted at the Fire Service College, a training center for emergency responders, where the system autonomously mapped three separate large industrial sites, which had a total ground coverage of $2528$ \SI{}{\meter\squared} and a maximum height of $27$~m. The sites were also mapped using manual pilot-flown missions and a Leica BLK360 \gls{tls} to provide comparative results. Quantitative and qualitative analysis of these results demonstrates the state-of-the-art autonomous mapping capabilities of Osprey as it captured maps of all the sites with equivalent or better coverage than the pilot-flown missions and the Leica BLK360 surveys. True colour reconstructions of the sites were also created from camera images captured by the Frontier sensor payload, using either a projection pipeline to colour the pointcloud map or SiLVR \citep{Tao2023}, a \gls{nerf} reconstruction approach.  

This paper is structured as follows: \tsecref{related} discusses related work on autonomous aerial mapping systems, with a focus on the different sensor configurations and mapping pipelines used. \tsecref{platform} presents the aerial platform used for Osprey, a DJI M600 with our custom sensor payload, Frontier, and the Rajant WiFi modules we used for communication. \tsecref{osprey} presents the Osprey mapping system, beginning with an overview of the pipeline and followed by a description of each component. \tsecref{experiments} presents the sites mapped during field experiments at the Fire Service College. \tsecref{evaluation} evaluates the mapping performance of Osprey on these sites in comparison to the pilot-flown missions and Leica BLK360 surveys. It also presents the true colour reconstructions. \tsecref{discussion} discusses the capabilities of Osprey and identifies directions for future work. \tsecref{conclusion} summarises the contributions of this work and lessons learned from developing Osprey.                                

\section{Related Work}
\seclabel{related}

Autonomous aerial mapping systems require five key components: (i) a sensor payload to perceive the environment, (ii) an odometry algorithm to estimate the platform motion, (iii) a mapping algorithm to aggregate sensor measurements into a combined representation, (iv) a mission planning algorithm to determine where measurements should be captured from and (v) a motion planning algorithm to identify safe paths through the environment. This section discusses related work on autonomous aerial mapping systems and their approaches for each component. The review focuses specifically on real world systems and excludes approaches that have only been demonstrated in simulation or motion capture environments.

\subsection{Sensor Payload}

Most aerial mapping systems use a sensor payload comprised of an \gls{imu} and either a camera and/or LiDAR. \gls{imu} measurements can provide high frequency estimates of the payload attitude while the visual or LiDAR measurements are used to obtain a more robust translation estimate and reconstruct the environment. Measurements from the different sensors used must be time synchronised either in hardware or software before being processed. The sensors used typically depend on the environment being mapped and the payload capacity of the aerial platform.

LiDAR-Inertial payloads \citep{Batinovic2021a,Hudson2022,Ohradzansky2022} are best suited for environments with distinctive geometric structures (e.g., large buildings) and environments with poor illumination or texture where visual perception may fail (e.g., underground). However, LiDARs typically weigh over $500$ grams and can not be lifted by smaller multirotor drones, which usually have a maximum payload capacity of a few hundred grams. Larger drones can carry these payloads and easily navigate around outdoor structures but may struggle to manoeuvre in narrow indoor or underground environments.  

Visual-Inertial payloads are suitable for highly textured and well illuminated environments where the platform can stay relatively close to the surfaces being mapped (e.g., indoors or in small outdoor scenes).  Many of these systems use a single stereo camera \citep{Bircher2016,Bircher2018a,Kompis2021a} mounted at a downwards angle, but some combine a stereo camera with a downward-facing monocular camera \citep{Tabib2022} or an optical flow sensor \citep{Heng2014} that is used solely for odometry. \citet{Teixeira2018} use a single monocular camera mounted at a downwards angle. These payloads are usually smaller and lighter than LiDAR-Inertial configurations, which allows them to be mounted on small aerial platforms that can navigate narrow passages more easily. Cameras are also typically less expensive than LiDARs, but multiple cameras are required to obtain the $360\degree$ horizontal field-of-view provided by most LiDARs.

Some systems combine measurements from all three sensing modalities as LiDAR-Visual-Inertial payloads. Earlier systems \citep{Shen2012,Yoder2016} use a 2D LiDAR and monocular camera configuration while more recent systems \citep{Papachristos2019,Nguyen2020,Scherer2022,Tranzatto2022a} use a 3D LiDAR and monocular camera. Using both LiDAR and visual sensors provides redundancy when operating in environments with both poorly textured surfaces and indistinct geometric structures where solely relying on either sensor may cause the odometry to fail. The disadvantage of these payloads is an increased weight, computational cost and the complexity of synchronising measurements from multiple sensors. A few systems use single-sensor payloads, either vision-only \citep{Song2020a,Song2021} or LiDAR-only \citep{Roucek2022}. These are simpler as no time synchronisation is required between sensors but are less robust to drift during rapid motions due to the lack of high frequency \gls{imu} measurements.  

The Osprey mapping system presented in this paper uses a custom LiDAR-Visual-Inertial sensor payload. This provides the sensor measurements necessary to obtain a robust multi-sensor odometry estimate and produce accurate true colour 3D maps of the target sites.   

\subsection{Odometry}

An odometry algorithm estimates the incremental platform motion using measurements from the sensor payload. Systems with LiDAR-Inertial payloads typically use odometry algorithms that register incoming LiDAR pointclouds to a rolling local map and use the \gls{imu} measurements to provide a high frequency motion estimate. \citet{Batinovic2021a} and \citet{Ohradzansky2022} use Google Cartographer \citep{Hess2016}, which obtains a motion estimate by scan matching LiDAR pointclouds with a local submap. This is created by aggregating pointclouds using initial motion estimates from the \gls{imu} measurements. \citet{Hudson2022} use odometry from Wildcat \citep{Ramezani2022a}, which performs surfel-based scan matching within sliding-window submaps and integrates \gls{imu} measurements in a continuous-time representation.  

Systems with Visual-Inertial payloads use either \gls{ekf}-based \citep[e.g., ROVIO;][]{Bloesch2015} or graph-based \citep[e.g., OKVIS or VINS-Mono;][]{Leutenegger2016,Qin2018} odometry algorithms to obtain a motion estimate from tightly coupled visual and \gls{imu} measurements. 

LiDAR-Visual-Inertial systems typically combine tightly coupled LiDAR-Inertial and Visual-Inertial estimators in a lightly coupled manner~\citep[e.g., Super Odometry or CompSLAM;][]{Zhao2021,Khattak2020}. Doing so retains the accuracy and resiliency of the tightly coupled motion estimates while exploiting the computational efficiency of a lightly coupled estimate. 

The vision-only system presented by \cite{Song2020a,Song2021} uses odometry from ORB-SLAM \citep{Mur-Artal2015a}. The LiDAR-only system from \citet{Roucek2022} uses odometry from A-LOAM, an extension of LOAM \citep{Zhang2014}, which obtains a motion estimate by extracting features from LiDAR pointclouds and matching them between scans.    

Osprey uses a multi-sensor fusion odometry algorithm to estimate the platform pose from \gls{imu} and LiDAR measurements \citep{Wisth2021b}. It provided reliable motion estimates at the large sites being mapped.

\subsection{Mapping}
A mapping algorithm aggregates the incremental sensor measurements captured from the moving platform into a map of the environment. Most systems use a volumetric representation in the form of a 3D voxel grid. The encoded information can include measurement occupancy, observation states, occlusion states and distances to other voxels. OctoMap \citep{Hornung2013} is a popular implementation which encodes the occupancy probability of each voxel; while Voxblox \citep{Oleynikova2017} encodes information on the distance to occupied voxels by integrating measurements into a \gls{tsdf} map representation. Some published approaches developed their own volumetric mapping algorithms. \cite{Shen2012} use a 2.5D occupancy grid map where each floor in an indoor environment is represented by a separate occupancy grid. \cite{Heng2014,Yoder2016} use custom implementations of a 3D occupancy grid and \cite{Scherer2022} present an implementation based on OpenVDB \citep{Museth2013}.

Volumetric representations are ideal for encoding the occupancy and exploration states of regions in an environment but the fidelity of the encoded surface information is constrained by the voxel grid resolution and using a high resolution typically incurs a high computational cost. \cite{Song2021} address this limitation by combining a volumetric representation with a surface representation that uses surfels to encode information about surface confidence. \cite{Tabib2022} use a novel \gls{gmm}-based map representation that combines elements of both volumetric and surface representations. \cite{Hudson2022} instead use a combined pointcloud and surfel representation. \cite{Roucek2022} use the pointcloud map representation provided by A-LOAM. \cite{Song2021} and \citet{Hudson2022} also incorporate pose graph updates from their SLAM algorithms (i.e., ORB-SLAM and Wildcat SLAM, respectively) into their map representations to correct odometry drift.

Osprey uses a graph-based SLAM algorithm to aggregate LiDAR pointclouds into a combined map and correct for odometry drift by detecting geometric loop closures \citep{Ramezani2020a}. A LiDAR-based localisation algorithm \citep{Kim2022} provides the ability to perform mapping over multiple flights.

\subsection{Mission Planning}                

A mission planning algorithm determines where the platform should capture measurements from in order to obtain a complete map of an environment. These algorithms typically adopt a \gls{nbv} planning approach which generates a set of potential views and then selects a next view from this set to obtain the best improvement in the current map. Some systems use view planning algorithms that generate potential views with sampling-based path planning algorithms. \cite{Bircher2016,Bircher2018a,Papachristos2019} sample views with \acrlong{rrt} \citep[\acrshort{rrt};][]{LaValle1998}\glsunset{rrt} and \cite{Song2020a} use RRT* \citep{Karaman2011}. This approach generates views that are well distributed within free space but does not explicitly consider the scene structure and therefore produces many views with limited value for extending the map. 

\cite{Nguyen2020} use GBPlanner \citep{Dang2019} and \cite{Tranzatto2022a} use GBPlanner2 \citep{Kulkarni2022}. These view planning algorithms sample views within a bounding box around the platform using \acrlong{rrg} \citep[\acrshort{rrg};][]{Karaman2011}\glsunset{rrg} and evaluate the value of these views for extending the map. GBPlanner2 extends GBPlanner by creating an adaptively sized bounding box from the local map instead of using a fixed size and only evaluating a subset of the sampled views.  

Many systems with volumetric representations use view planning algorithms that generate potential views directly from specific voxel types, which helps to ensure their utility for extending the map. These can be occupied voxels \citep{Yoder2016,Teixeira2018}, surface frontier voxels \citep[i.e., occupied voxels with unknown neighbours;][]{Kompis2021a,Scherer2022} or exploration frontier voxels \citep[i.e., free voxels with unknown neighbours;][]{Heng2014,Song2021,Batinovic2021a,Roucek2022,Ohradzansky2022}. Systems that do not maintain a volumetric representation use other view generation methods. \citet{Tabib2022} generate views using motion primitives and \citet{Hudson2022} propose views by identifying point-based frontiers \citep{Williams2020a}.

Once a set of potential views is generated a particular next best view is selected from the set based on a utility metric. Some methods simply select the view closest to the platform \citep{Heng2014,Roucek2022} or one that lies a given inspection direction \citep{Teixeira2018}. Most systems with volumetric map representations attempt to maximise coverage of specific voxel types relative to the travel cost of reaching a view. These can be unknown voxels \citep{Bircher2016,Bircher2018a,Papachristos2019,Tranzatto2022a}, surface frontier voxels \citep{Yoder2016,Kompis2021a,Scherer2022} or exploration frontier voxels \citep{Song2020a,Song2021,Batinovic2021a,Ohradzansky2022}. Some view selection metrics also consider surface quality \citep{Kompis2021a} or platform momentum \citep{Scherer2022}. \citet{Song2020a,Song2021} extend the selection of a next best view to a next best trajectory by optimising the path to a chosen view to maximise coverage of exploration frontier voxels \citep{Song2020a} and low confidence surfels \citep{Song2021}. \citet{Tabib2022} select a motion primitive to reduce map uncertainty and \citet{Hudson2022} select a view based on reachability metrics.

Mission planning for Osprey is performed by a \gls{nbv} planning algorithm that operates directly on the LiDAR measurements \citep{Border2022a}. Measurement-based frontiers are identified by performing a density-based classification of each measurement. Potential views are generated to observe these frontiers and a next best view is chosen to observe the most frontiers per unit of travel distance. 

\subsection{Motion Planning}

Motion planning is a lower level task which identifies a specific collision-free path for the platform to traverse through the environment to reach a view selected by the mission planner. Systems that use sampling-based path planners to generate potential views typically use the collision-free path to the chosen view already provided by the existing planning graph. Meanwhile, systems with voxel-based view generation methods often create their own planning graphs by connecting view-based vertices with collision-free edges and find a path through the graph using A* search \citep{Hart1968} or Dijkstra's algorithm \citep{Dijkstra1959}. \cite{Yoder2016} use the SPARTAN algorithm \citep{Cover2013}, \cite{Teixeira2018} use MISP \citep{Alzugaray2017} and \cite{Roucek2022} use a spatial exploration approach \citep{Bayer2019}. Other systems use local planning methods including vector fields \citep{Heng2014}, direct vision-based control \citep{Ohradzansky2022} and motion primitives \citep{Kompis2021a,Tabib2022}. \citet{Scherer2022} plan paths with RRT-Connect \citep{Kuffner2000}, \cite{Batinovic2021a} use \gls{rrt}, \cite{Shen2012,Song2021} use RRT* and \citet{Hudson2022} use a motion planner from Emesent. 

Osprey uses a sampling-based planner that can quickly identify collision-free paths for the platform, often in less time than \gls{rrt} or RRT*, and continuously improves its solution within the remaining planning time \citep{Strub2022}.

\section{Aerial Platform}
\seclabel{platform}

The Osprey aerial platform \figref{sensor-plat} is a DJI M600 drone with a custom sensor payload, called \emph{Frontier}\footnote{More details on the payload can be found here: \url{https://ori.ox.ac.uk/labs/drs/frontier-payload/}}, that was developed at the \gls{ori}. A Rajant WiFi module provides wireless communication between the platform and an operator. The DJI M600 is a hexrotor drone with an integrated flight controller, \gls{imu} and GPS. It has a flight time of up to $30$ minutes when carrying the $3\,$kg payload, which includes the Frontier, a Rajant module and an external battery. 

The Frontier is comprised of a Hesai QT64 LiDAR (\tfigref{sensor-plat}a), a Sevensense Core Research sensor module (\tfigref{sensor-plat}b) with a Bosch BMI085 IMU and Sony IMX-273 colour fisheye cameras on three sides, and an Intel NUC computer with a 12-core Intel i7 CPU and 32GB of RAM  (\tfigref{sensor-plat}c). \ttblref{sensors} presents the specifications for these sensors. The wide vertical field-of-view of the Hesai LiDAR allows the platform to fly above tall structures while still retaining enough visibility of surfaces below it for good odometry and mapping performance. The fisheye cameras were not used for the online mapping system but their images were recorded and used to produce colour reconstructions offline.

The camera intrinsics and camera-IMU extrinsics were calibrated with Kalibr \citep{kalibr}. The camera-LiDAR extrinsics were calibrated with DiffCal \citep{diffcal}. Extrinsics between the Frontier and DJI base frames were manually calibrated. This transformation was necessary when sending commands from the Frontier to the DJI flight controller. The sensors are all connected to the NUC via Ethernet and time synchronised with PTP. All processing for the system is performed onboard the drone using the NUC.

Communication between the platform and an operator was achieved using Rajant WiFi modules, which create a reconfigurable wireless mesh network and allowed reliable communication to be maintained between the platform and operator when mapping large sites. One Rajant BreadCrumb ES1 module (\tfigref{sensor-plat}d) was mounted on the aerial platform, another was connected to the operator laptop and intermediate connectivity was provided by Rajant BreadCrumb DX2 modules (\tfigref{sensor-plat}e) distributed around the target site. Connectivity was required to send high level commands (e.g., to start or stop a mission) to the platform and monitor the system for safety but was not critical for the autonomous mapping, which ran entirely on the Frontier.   

\begin{figure}[tp]
	\centering
	\captionsetup[subfigure]{labelformat=empty}
	\subfloat[]{\includegraphics[width=0.75\linewidth]{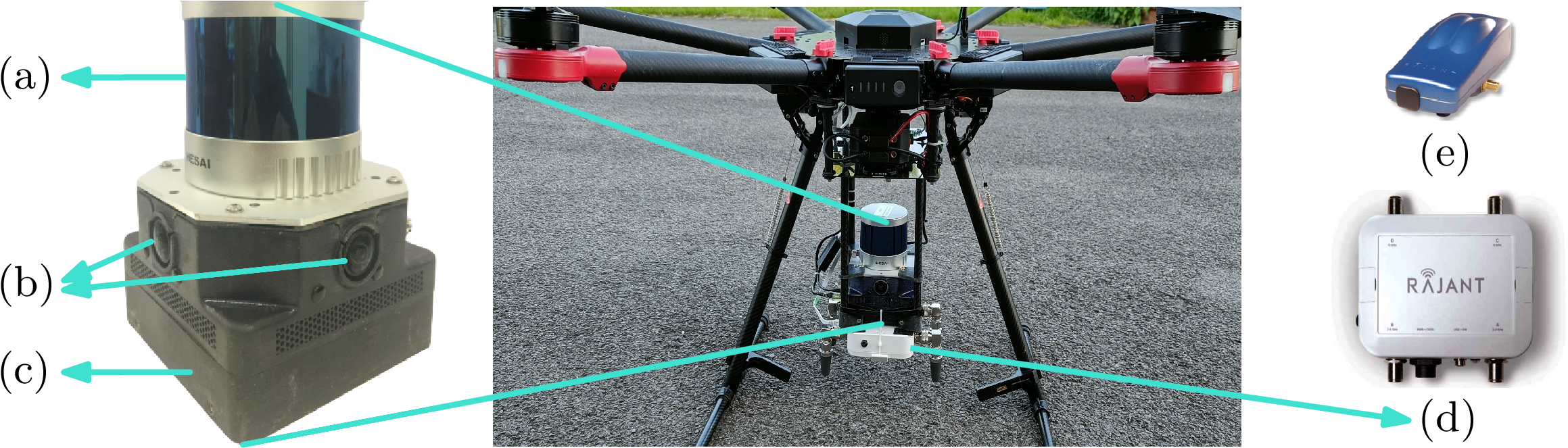}}
	\vspace{-4ex}
	\caption{The Osprey aerial platform, a DJI M600 drone with a Frontier and Rajant WiFi module mounted underneath it. The Frontier consists of (a) a Hesai QT64 LiDAR, (b) a SevenSense Core Research sensor module with an IMU and three 1.6 megapixel colour fisheye cameras, and (c) an Intel NUC computer. Communication with the operator is provided by a wireless mesh network comprised of (d) a Rajant BreadCrumb ES1 module on Osprey and (e) Rajant BreadCrumb DX2 modules distributed around a target site.}	
	\figlabel{sensor-plat}
	\vspace{-2ex}
\end{figure}

\begin{table}[tp]
	\centering
	\caption{Specifications for the LiDAR, IMU and cameras used by the custom Frontier sensor payload.}
	\vspace{1ex}
	\tbllabel{sensors}
	\begin{adjustbox}{width=0.8\linewidth,center}
		\begin{tabular}{@{}llll@{}}
			\toprule
			& Hesai QT64 LiDAR & Sony IMX-273 Cameras & Bosch BMI085 IMU \\ \midrule
			Horizontal Field-of-View [\degree]       & 360              & 126                  & ---              \\
			Vertical Field-of-View [\degree]       & 104.2            & 92.4                 & ---              \\
			Horizontal Resolution [pt or px] & 600              & 1440                 & ---              \\
			Vertical Resolution [pt or px]   & 64               & 1080                 & ---              \\
			Frequency (Hz)                & 10               & 20                   & 400              \\ \bottomrule
		\end{tabular}
	\end{adjustbox}
\end{table} 

\section{Osprey Mapping System}
\seclabel{osprey}

The Osprey mapping pipeline \figref{osprey-pipeline} can be divided into five key components: (i) Sensor Payload, (ii) Odometry, (iii) Mapping, (iv) Mission Planning and (v) Motion Planning. The Sensor Payload captures measurements from the \gls{imu}, LiDAR and cameras. 

In the Odometry component, the \gls{vilens} algorithm \citep{Wisth2021b} obtains a robust motion estimate by tightly fusing \gls{imu} and LiDAR measurements in a factor graph. Motion undistortion is also applied to the raw LiDAR measurements, which are then published as processed pointclouds. These are integrated into an OctoMap representation of the environment, which is used for collision checking by the motion planning algorithm. 

\begin{figure}[tp]
	\centering
	\captionsetup[subfigure]{labelformat=empty}
	\subfloat[]{\includegraphics[width=\linewidth]{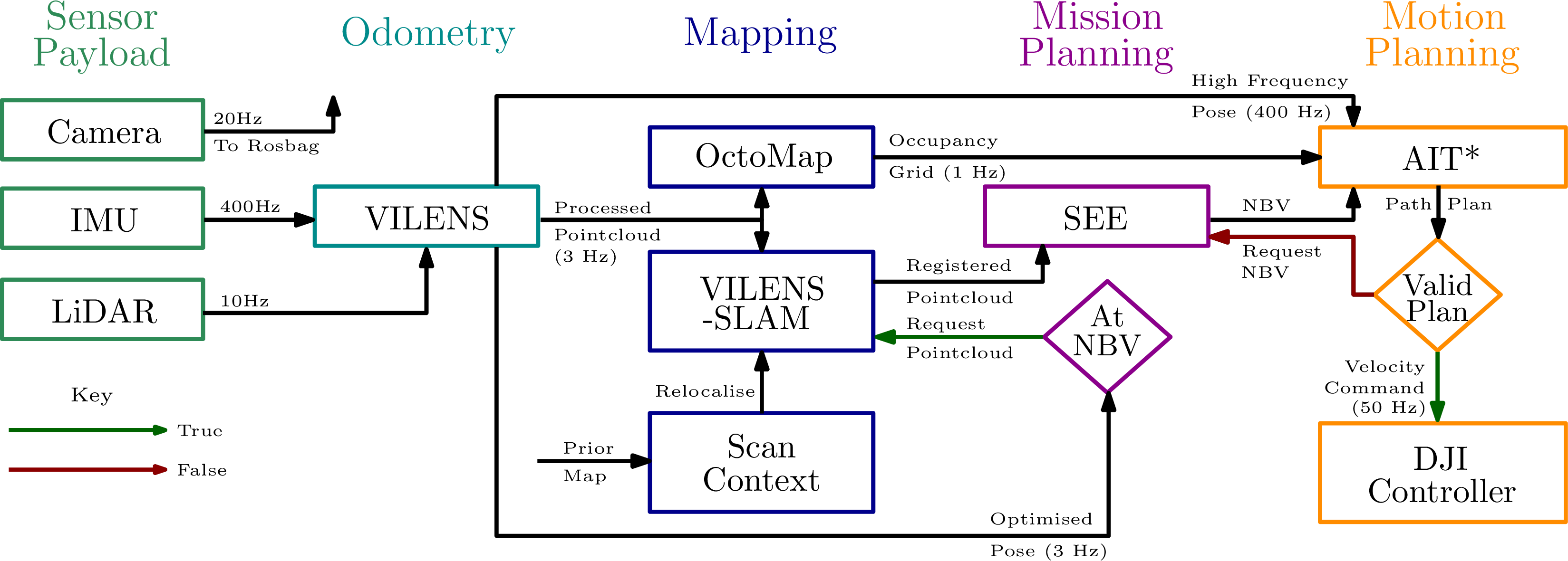}}
	\vspace{-4ex}
	\caption{Flowchart showing an overview of the five key components of the Osprey mapping pipeline: Sensor Payload, Odometry, Mapping, Mission Planning and Motion Planning. The rectangular boxes represent sensor inputs or algorithms and the connections between them denote data pipelines. The diamonds represent boolean decisions, with green arrows denoting true outcomes and red arrows denoting false outcomes.}	
	\figlabel{osprey-pipeline}
	\vspace{-2ex}
\end{figure}

In the Mapping component, the VILENS-SLAM  algorithm \citep{Ramezani2020a} integrates undistorted LiDAR pointclouds from VILENS into a pose graph and detects geometric loop closures to correct for odometry drift. A combined map is created by aggregating every registered pointcloud in the pose graph in a common reference frame. The most recent registered cloud and pose graph are continuously published. An implementation of ScanContext \citep{Kim2022} provides the relocalisation capabilities necessary to map large areas over multiple flights.

In the Mission Planning component, the Surface Edge Explorer \citep[SEE;][]{Border2022a}\glsunset{see}, a \gls{nbv} planning algorithm, directs the platform to capture measurements from views around a target site with the aim of obtaining complete coverage of visible surfaces. It processes the pointclouds captured at these chosen views and adds the measurements to its internal map representation. Boundaries between complete and incomplete surfaces are identified and potential views are generated to extend coverage of the map. A next best view is iteratively selected from this set of potential views and sent to the motion planning algorithm.

In the Motion Planning component, the \acrlong{aitstar} \citep[\acrshort{aitstar};][]{Strub2022}\glsunset{aitstar} algorithm plans a collision-free path for the platform to traverse between its current pose and the next best view chosen by \gls{see}. It uses the OctoMap occupancy grid for collision checking. If \gls{aitstar} finds a valid path then velocity commands are generated for the platform and sent to the DJI flight controller. If a valid path is not found then \gls{aitstar} requests an alternative view from \gls{see}. 

The following subsections provide detailed descriptions of the algorithms used in each component of the mapping pipeline. The values used for the key parameters of each algorithm are presented in \tsecref{params}. 

\subsection{Sensor Payload}

Measurements are captured from the LiDAR at $10$~Hz, the \gls{imu} at $400$~Hz and the cameras at $20$~Hz. All of the sensors are synchronised with the onboard computer via PTP to ensure millisecond accurate time alignment between the measurements, which is critical for reliable operation of the \gls{vilens} odometry algorithm. The LiDAR and \gls{imu} measurements are sent to VILENS for processing but the camera images are only used for post-processed colour reconstructions and are saved directly to a rosbag.    

\subsection{Odometry}

The odometry algorithm incrementally estimates the platform pose as it moves around a target site. This estimate is computed from the sensor measurements and provides the initial pose for LiDAR pointclouds that are combined by the mapping algorithm. The Osprey system uses \gls{vilens} \citep{Wisth2021b}, a multisensor fusion approach that obtains a robust and reliable odometry estimate by integrating tightly coupled measurements from different sensors into a factor graph with GTSAM\footnote{\url{https://gtsam.org/}} \citep{gtsam}. The DJI flight controller separately maintains an estimate of its own state using onboard IMU and GPS sensors but this is not used for odometry or mapping as it is less accurate than \gls{vilens} and can degrade significantly when the platform has limited GPS connectivity (e.g., when flying close to large structures). 

\begin{figure}[tp]
	\centering
	\captionsetup[subfigure]{labelformat=empty}
	\subfloat[]{\includegraphics[width=0.4\linewidth]{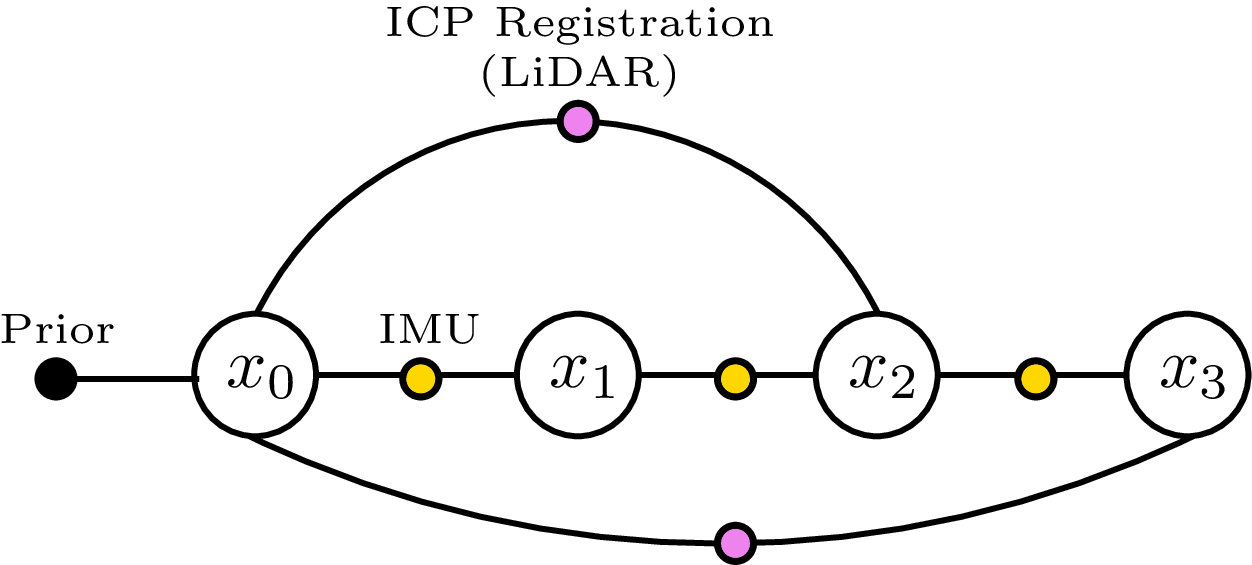}}
	\vspace{-3ex}
	\caption{An illustration of the \gls{vilens} factor graph, showing the states, $x_i$, and the factors connecting them from the \gls{imu} (yellow) and \gls{icp} registration (pink) modules.}	
	\figlabel{odometry}
\end{figure}

\gls{vilens} uses a modular approach to integrate measurements from different sensors. In the large structured sites mapped by Osprey the best results were obtained using the \gls{imu} and \gls{icp} registration modules. The \gls{imu} module provides the core of the factor graph by pre-integrating \gls{imu} measurements between consecutive states at the full frequency of the \gls{imu} \figref{odometry}. This provides an initial constraint on the motion between states and produces a high frequency pose estimate that can be used to generate control commands (e.g., for the DJI flight controller). The IMU pose estimates are also used to apply motion undistortion to the LiDAR pointclouds, which improves the performance of the \gls{icp} registration module.

The \gls{icp} module uses a scan-to-submap registration approach. This typically produces a more robust transformation estimate than scan-to-scan registration as a submap provides more structural detail than a single pointcloud. The submap is created by aggregating pointclouds captured from poses within $25$~m of the current pose into a local map. The map density is constrained by enforcing a minimum separation distance between the included poses and downsampling the final map to $30000$ points. The submap is divided into a \emph{dense} region, within $4$~m of the current pose, where the separation distance is $0.25$~m and a \emph{sparse} region where the separation distance is $2$~m. This ensures that more detailed information is retained for nearby structures and improves the \gls{icp} accuracy without significantly increasing the computational cost.    

Incoming pointclouds are also downsampled to $30000$ points and then registered against the submap to obtain transformation estimates. This typically occurs at a lower frequency than the LiDAR output due to the computational cost of the \gls{icp} registration. \gls{icp} registration factors are defined by the relative transformation of a registered pointcloud to the current submap reference frame and connect their corresponding non-consecutive states in the factor graph \figref{odometry}. The factor graph is solved after adding each \gls{icp} factor to provide an optimised pose estimate at the frequency of the registration module. 
\subsection{Mapping}

The mapping algorithm aggregates sensor measurements captured as the platform moves through the environment into a combined map. The pose estimate provided by the odometry algorithm is used to provide an initial alignment of these measurements in a common reference frame. The Osprey system uses VILENS-SLAM \citep{Ramezani2020a}, a pose-graph based mapping algorithm with loop closure capabilities.

The VILENS-SLAM pose graph \figref{mapping} is implemented as a factor graph using GTSAM. Nodes in the graph represent LiDAR pointclouds captured at different poses. Edges between consecutive nodes represent odometry factors, which are provided by pose estimates from VILENS. New pointclouds are added to the pose graph at a set distance interval (e.g., every $1$~m) or when requested by the mission planning algorithm. They are processed by VILENS for motion undistortion before being added to the pose graph, which improves the accuracy of the captured map.

\begin{figure}[tp]
	\centering
	\captionsetup[subfigure]{labelformat=empty}
	\subfloat[]{\includegraphics[width=0.7\linewidth]{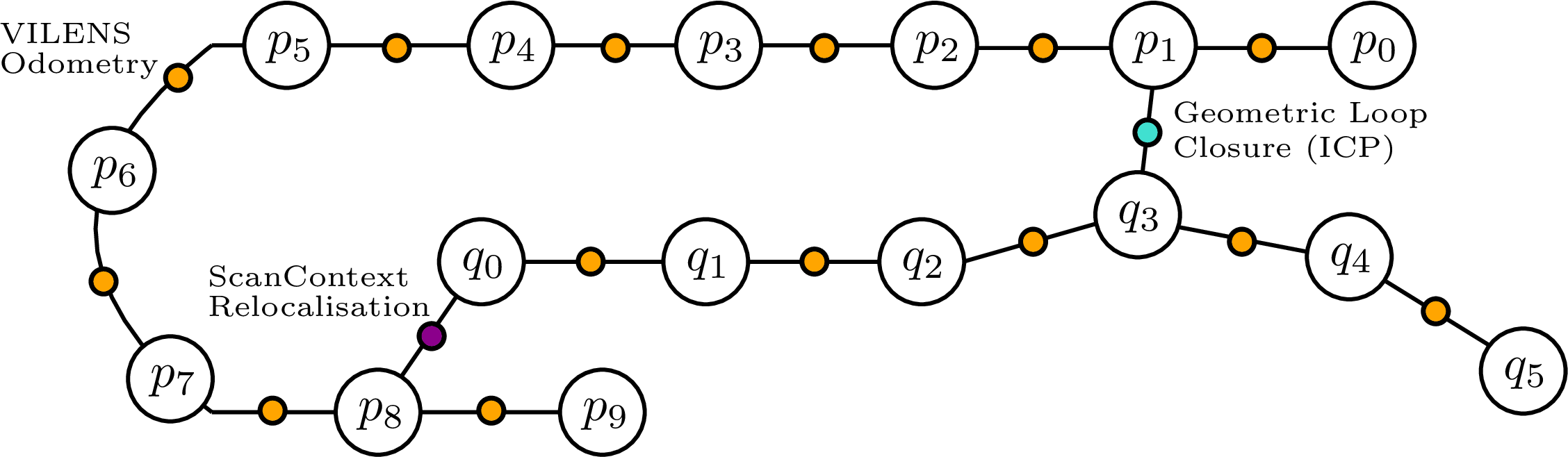}}
	\vspace{-3ex}
	\caption{An illustration of a VILENS-SLAM pose graph created from a multi-session mapping mission. Nodes associated with the first session are denoted as $p_i$ and those from a subsequent session are denoted as $q_i$. Odometry factors from VILENS (orange) connect consecutive nodes in the same session. Geometric loop closure factors (light blue) can exist between nodes in the same or different sessions, but only an inter-session loop closure is shown for simplicity. A relocalisation factor (purple) is added between nodes in different sessions after ScanContext identifies a successful match.}	
	\figlabel{mapping}
\end{figure}

VILENS-SLAM uses geometric loop closures to correct for odometry drift in the map. Loop closure detection is proximity-based and occurs when the pose of a newly added pointcloud is within a given radius of existing poses in the graph. When a loop closure candidate between two poses is identified an \gls{icp} registration is performed between their associated pointclouds. If this registration is successful then the loop closure is considered valid and the computed transformation between the two pointclouds is added to the graph as a loop closure factor \figref{mapping}. Registration success is determined by convergence of the transformation and the number of inliers in the final result. The factor graph is then optimised to account for the new factor and the poses of the pointclouds in the pose graph are updated. The transformation between the VILENS odometry and VILENS-SLAM map frames is also updated to correct for drift in the VILENS pose estimate.

\subsubsection{Multi-session Mapping}
\seclabel{multimap}

Our system can extend pose graphs across multiple sessions in order to map large sites. This requires the ability to relocalise the platform within an existing map. VILENS-SLAM uses an implementation of ScanContext \citep{Kim2022} to perform place recognition between new LiDAR pointclouds and those integrated in an existing pose graph. ScanContext creates a unique descriptor for each pointcloud by segmenting the cloud into bins defined by circular sectors and rings, which are centered on the pointcloud origin. The descriptor is defined by the greatest $z$-axis value of measurements within each bin. This descriptor is robust to rotational variations around the $z$-axis, which allows the system to relocalise when the yaw orientation of the aerial platform differs between visits to the same location.

The extension of an existing map when starting a new flight as part of an ongoing mission proceeds as follows. The existing SLAM map and pose graph are loaded from files and ScanContext descriptors are computed for all the individual pointclouds. Descriptors generated for online pointclouds from the LiDAR are matched with those from the pointclouds in the pose graph. When a valid match is found an \gls{icp} registration is performed between the matched pointclouds to verify the match and obtain a transformation estimate. If the \gls{icp} registration converges to a good solution then the relocalisation is successful. The online pointcloud is added to the pose graph and connected to the node associated with its matched pointcloud from the existing map by a relocalisation factor \figref{mapping}, which uses the transformation obtained from the \gls{icp} registration. The integration of new pointclouds into the map then proceeds as previously described.        

\subsection{Mission Planning}

The mission planning algorithm determines where measurements should be captured from in order to improve map coverage. Osprey uses \gls{see} \citep{Border2022a}, a measurement-direct \gls{nbv} approach that plans views to obtain a minimum measurement density from all surfaces within a bounded volume of space. The platform is placed facing a target structure and the bounding volume is then defined by the user relative to this takeoff location as an axis-aligned bounding box that is large enough to encompass the structure.  

\gls{see} starts planning potential views after takeoff by capturing and processing a registered LiDAR pointcloud from VILENS-SLAM. The pointcloud measurements are integrated into the unstructured pointcloud representation used by \gls{see} and individually classified based on the local density of neighbouring points within a resolution radius. Measurements with a minimum density are classified as \emph{core} points and represent complete surfaces. Measurements with insufficient density are \emph{outliers} and denote incomplete surfaces. The boundary between complete and incomplete surfaces is defined by \emph{frontier} points, which are outlier points with core point neighbours \figref{mission-planning}.

\begin{figure}[tp]
	\centering
	\captionsetup[subfigure]{labelformat=empty}
	\subfloat[]{\includegraphics[width=0.55\linewidth]{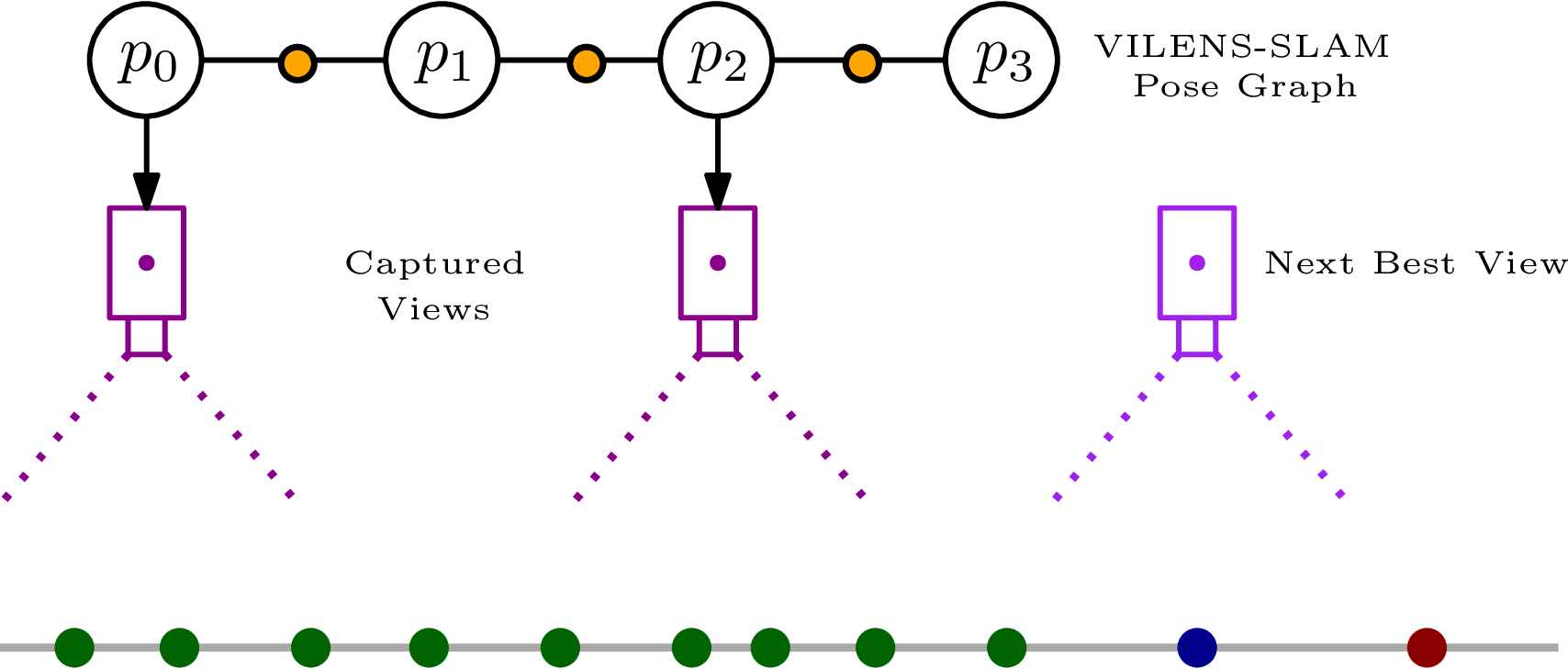}}
	\vspace{-3ex}
	\caption{An illustration of how SEE, the mission planning algorithm, integrates with the pose graph created by VILENS-SLAM, the mapping algorithm. Views captured by SEE are denoted by sensor icons (dark purple) and correspond to reference clouds in the pose graph. Measurements processed by SEE lie on a surface (grey line) and are classified as core (green dots), frontier (blue dot) or outlier (red dot) points. The next best view chosen to capture the surface around the frontier is shown as a sensor icon (light purple).}	
	\figlabel{mission-planning}
\end{figure}

Potential views are generated from the frontier points to extend the map coverage by capturing new measurements at the boundary between complete and incomplete surfaces. Each view is generated to observe the surface around its associated frontier point at a specified distance and with an orientation orthogonal to the local surface in order to maximise coverage. If a frontier point is occluded from its associated view then an optimisation approach is used to find an alternative unoccluded view. 

\gls{see} iteratively chooses a next best view from the set of potential views to extend map coverage. This next best view is chosen to provide the greatest improvement in surface coverage per unit of travel distance. The chosen view is sent to the motion planning pipeline, which identifies a collision-free path and commands the platform to reach the view. When the next best view is reached the platform stops and \gls{see} captures a new registered LiDAR pointcloud from VILENS-SLAM for processing. \gls{see} continues to plan new potential views and select next best views until all measurements are classified as either core or outlier points.        

The version of \gls{see} used for Osprey extends our previous work \citep{Border2022a} by making the point classification more robust to measurement noise, adding the ability to integrate loop closures and enabling multi-session mission planning. The misclassification of points on complex surfaces (e.g., thin or pointed structures) is reduced by evaluating the distribution of the neighbouring points around a measurement in addition to their density. \gls{see} integrates loop closures from VILENS-SLAM into its map representation by receiving pose graph updates when a loop closure is added and computing the transformation between the previous and current poses of each pointcloud in its internal map. The corresponding measurements in the pointclouds are then transformed and reprocessed by \gls{see} as if they were newly added. 

Multi-session mission planning capabilities were added to \gls{see} by enabling it to export its current observation state at the end of a flight and import a previous state when starting a new flight as part of an ongoing mission. When beginning a new flight that will extend an existing map, \gls{see} is started with the same parameters as the last flight and imports the previous observation state. It waits for ScanContext to relocalise against the existing map and then selects a next best view from the imported observation state.

\subsection{Motion Planning}

The motion planning algorithm identifies a collision-free path for the platform to follow from its current pose to the selected next best view \figref{motion-planning}. Osprey uses \gls{aitstar} \citep{Strub2022}, an almost-surely asymptotically optimal sampling-based planner, to efficiently plan trajectories between views. \gls{aitstar} uses informed sampling to obtain an approximation of the state space solely within a constrained region where an improvement to the current path could be found. It is able to efficiently evaluate new paths through the informed samples by simultaneously calculating and exploiting a problem-specific cost heuristic that aims to minimise the path length. It finds an initial path first and then continuously refines its informed state approximation and cost heuristic to identify more efficient paths within the remaining allocated planning time. This enables \gls{aitstar} to find efficient paths faster than other sampling-based planners (e.g., \gls{rrt}, RRT*).

\begin{figure}[tp]
	\centering
	\captionsetup[subfigure]{labelformat=empty}
	\subfloat[]{\includegraphics[width=\linewidth]{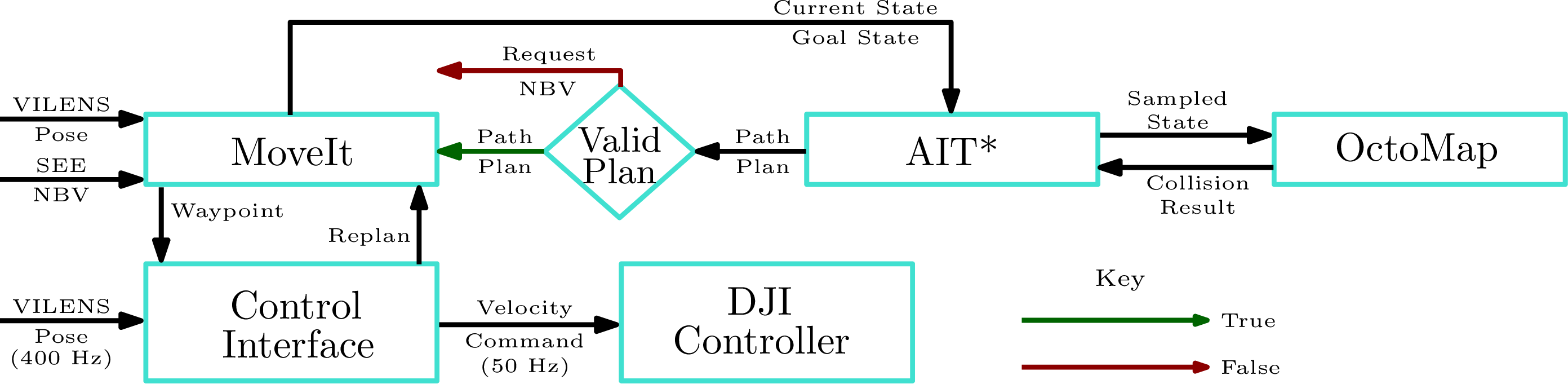}}
	\vspace{-3ex}
	\caption{A flowchart overview of the motion planning pipeline. The constituent algorithms are represented by rectangular boxes and the connections between them represent data communications. Diamond boxes denote decisions, with green arrows indicating true outcomes and red arrows indicating false outcomes. }	
	\figlabel{motion-planning}
\end{figure}

Osprey uses the implementation of \gls{aitstar} provided with the Open Motion Planning Library \citep[OMPL\footnote{\url{https://ompl.kavrakilab.org/}};][]{ompl} and uses MoveIt\footnote{\url{https://moveit.ros.org/}} \citep{Coleman2014} to interface between AIT* and the DJI M600 flight controller. Collision checking is performed using an OctoMap occupancy grid, which has a $0.5$~m voxel resolution and is updated with registered pointclouds from VILENS-SLAM at $1$~Hz. 

The planning pipeline works as follows. MoveIt receives a next best view from \gls{see} and the current platform pose from \gls{vilens}. These $6$~DoF poses are transformed into $4$~DoF states consisting of a position and yaw angle. \gls{aitstar} then attempts to plan a path between the start state (i.e., the current platform state) and the goal state (i.e., the next best view state) within a maximum allowed planning time of $10$~s. 

If a valid path is not found (e.g., because the goal state is too close to an obstacle) then an alternative view is requested from \gls{see}. If a valid path is found then the first state along the trajectory is sent to the control interface, which generates velocity commands for the DJI flight controller. The platform stops when each state is reached (i.e., the VILENS pose is within $1$~m of the target state) and a new path to the goal is planned from the current state. This helps prevent collisions with previously unobserved obstacles. The platform continues moving between states and replanning its trajectory until the goal state is reached.

The control interface generates velocity commands for the DJI flight controller at $50$~Hz based on the difference between the current platform state, which is updated from the high frequency \gls{vilens} pose at $400$~Hz, and the next target state. The platform is commanded to accelerate at constant rate, $a$, until the maximum allowed velocity, $v$, is reached or it nears the target state and begins decelerating to a stop.

\begin{figure}[!tp]
	\centering
	\captionsetup[subfigure]{labelformat=empty}
	\subfloat[]{\includegraphics[width=\linewidth]{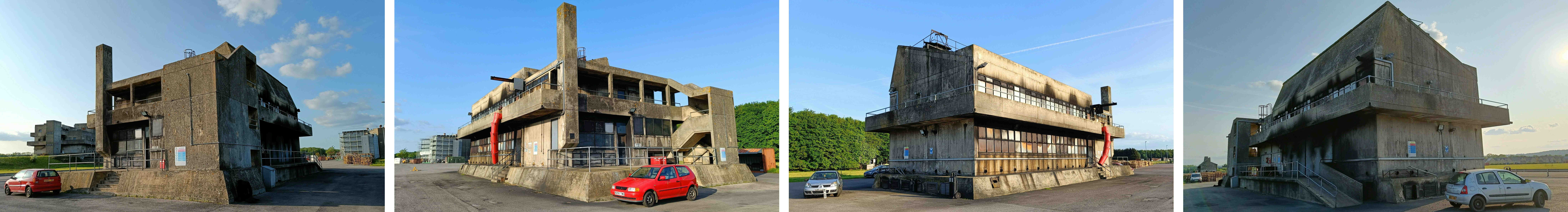}}
	\vspace{-4ex}
	\caption{Photographs of the \emph{Industrial Building A} site. It is a two storey building, 30x16x12m (LxWxH) in size, with a balcony surrounding the upper storey, which is partially enclosed on one side. There is a concrete fire escape on one corner of the building, a garbage chute (red) attached to one side, a concrete chimney and various metal structures on the roof.}
	\figlabel{indy_b_photos}
\end{figure}

\begin{figure}[!tp]
	\centering
	\captionsetup[subfigure]{labelformat=empty}
	\subfloat[]{\includegraphics[width=0.7\linewidth]{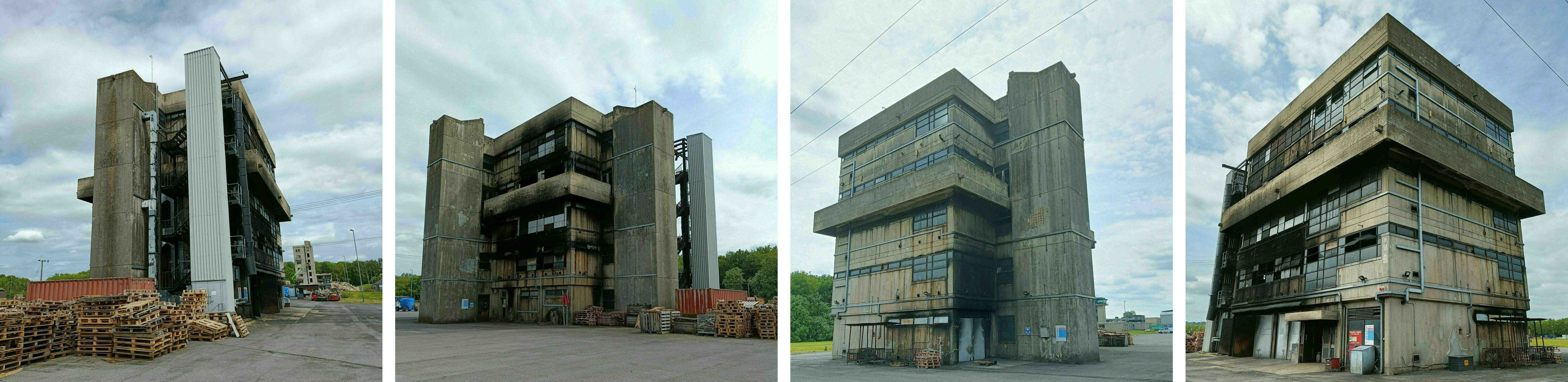}}
	\vspace{-4ex}
	\caption{Photographs of the \emph{Industrial Building B} site. It is a five storey building, 27x20x21m (LxWxH) in size, with a balcony surrounding the fourth storey and two concrete fire escapes attached to different sides of the building. There is also an exposed metal fire escape with a garbage chute attached to the building, which is comprised of thin metal structures and proved challenging to map.}
	\figlabel{indy_a_photos}
\end{figure}

\begin{figure}[!tp]
	\centering
	\captionsetup[subfigure]{labelformat=empty}
	\subfloat[]{\includegraphics[width=\linewidth]{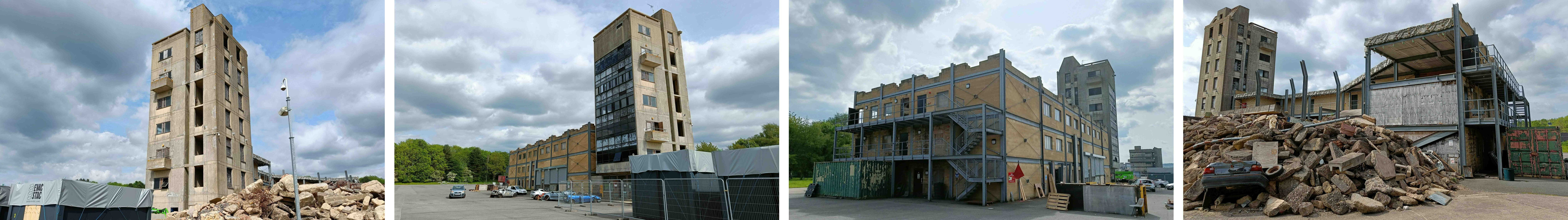}}
	\vspace{-4ex}
	\caption{Photographs of the \emph{Rig 5} site. It is a large site, 52x29x27m (LxWxH) in size, which contains multiple structures and is staged as a disaster scenario. The site is comprised of a seven storey building next to a long three storey building with an exposed interior on one side and a metal fire escape at one end. There is a large rubble pile next to these buildings with several upright metal girders within it.}
	\figlabel{rig_5_photos}
\end{figure}

\section{Field Experiments}
\seclabel{experiments}

Field experiments with Osprey were conducted at the Fire Service College, Moreton-in-the-Marsh, Gloucestershire, UK. The facility is a training center for firefighters with several standalone buildings ideal for testing the Osprey autonomous mapping system. Field experiments were conducted at three of these sites ranging from a medium-sized two storey industrial building (\emph{Industrial Building A}; \tfigref{indy_b_photos}) to a large five storey industrial building with several external fire escapes (\emph{Industrial Building B}; \tfigref{indy_a_photos}) and an even larger site containing multiple structures that was staged as a disaster scenario (\emph{Rig 5}; \tfigref{rig_5_photos}).

\subsection{Experiments}          

Three types of experiment were performed at each site: a survey-grade scan using a \gls{tls}, a pilot-flown mission and an Osprey autonomous mapping mission. A Leica BLK360 was used to obtain a survey-grade map of each site by registering individual scans around the site. The pilot-flown missions were performed by an author who is an experienced pilot and flew the platform around each site while the SLAM algorithms ran passively onboard. The pilot empirically judged when the map was complete. These experiments were performed to provide comparative results for evaluating the performance of the Osprey system.

The Osprey missions were conducted with an operator overseeing progress on a laptop and a safety pilot monitoring the aerial platform, ready to take control if necessary. These autonomy missions proceeded as follows. The aerial platform was placed at one end of a site and a bounding box, which defined the target area for mission planning, was specified by the operator. The bounding box was set large enough to encompass the entire site, with a minimum height of $2$~m and a maximum height greater than the tallest structure. 

An autonomous mission was initiated by the operator using the automatic takeoff capability of the DJI M600. After takeoff, the platform would climb to a height of $3$~m and the autonomous mapping system would start. The platform would then be directed by the Osprey system to fly autonomously around the site. The safety pilot monitored the platform continuously and would intervene if there appeared to be a risk of collision. Interventions only occurred on a few occasions when the drone deviated from its planned trajectory due to wind gusts or poor performance of the flight controller (e.g., due to degraded GPS accuracy). In these cases the safety pilot would take control, fly the platform to a safe minimum distance of at least $3$~m from any structure and wait until the gust passed or the flight controller performance improved (e.g., due to better GPS accuracy). Control was then returned to the Osprey system and the autonomous mission was resumed. When the platform battery level was insufficient to continue flying the operator would stop the autonomous mission planning and manually land the platform.         
        
After landing, the platform batteries would be replaced and a new flight could be initiated to extend the current map as part of the same autonomous mission. At the start of this next flight the Osprey system would load the current map from file and relocalise within it using ScanContext before activating the autonomous mission planning. This relocalisation would typically occur during takeoff, but in some cases the safely pilot needed to manually fly the platform around the takeoff area until the relocalisation was successful. The mission planning would then be started. When the mission planning algorithm determined that the target area had been completely mapped it would stop and return control to the safety pilot for a manual landing.           

\subsection{Parameters}
\seclabel{params}

The algorithms used the same parameters in all of the Osprey experiments, with only the bounding box being site-specific. The key parameters for each algorithm are presented in \ttblref{parameters-table}. \gls{vilens}, the  odometry algorithm, ran its \gls{icp} module on every third pointcloud from the LiDAR (i.e., at a frequency of $3$~Hz). The scan-to-submap \gls{icp} registration was successful if the difference in translation between iterations converged to less than $0.01$~m and the difference in rotation converged to less than $0.001$~rad within $30$ iterations.   

VILENS-SLAM, the mapping algorithm, integrated new pointclouds from the LiDAR into its pose graph every $1$~m, which provided sufficient map density while maintaining a reasonable computational cost. The detection radius for geometric loop closures between nodes in the pose graph was $3$~m, which was large enough to correct for odometry drift while mitigating the risk of false loop closure matches. A geometric loop closure was considered successful if the \gls{icp} registration met the same convergence criteria used by VILENS and there were at least $5000$ inliers in the final \gls{icp} result. 

\gls{see}, the mission planning algorithm, used a view distance of $10$~m for potential views in order to obtain high coverage of the target site from each captured view. The resolution radius of $1.5$~m and target density of $5$ points per m$^3$ were chosen to ensure that the boundaries between complete and incomplete surfaces could be reliably identified on the large structures being observed while mitigating the effects of measurement noise. 

The motion planning pipeline, which is built around \gls{aitstar}, used a maximum distance of $5$~m between waypoints to ensure the platform did not attempt to traverse unobserved regions without replanning its path. The maximum velocity of $1$~m per s and acceleration of $0.1$~m per s$^2$ were chosen based on limitations of the DJI flight controller and to ensure that the safety pilot could take control of the platform if required.     

\renewcommand{\arraystretch}{1.1}
\begin{table}[]
	\centering
	\caption{The key parameters used by each of the constituent algorithms in the Osprey mapping system.}
	\vspace{1ex}
	\tbllabel{parameters-table}
	\begin{adjustbox}{width=\linewidth,center}
		\aboverulesep=0ex
		\belowrulesep=0ex
		\begin{tabular}{@{}lclllclc@{}}
			\toprule
			\multicolumn{2}{c}{Odometry (VILENS)} &
			\multicolumn{2}{c}{Mapping (VILENS-SLAM)} &
			\multicolumn{2}{c}{Mission Planning (SEE)} &
			\multicolumn{2}{c}{Motion Planning (AIT*)} \\ \midrule
			ICP Module Freq. [Hz] &
			\multicolumn{1}{c|}{3} &
			Reference Cloud Dist. [m] &
			\multicolumn{1}{c|}{1} &
			View Dist. [m] &
			\multicolumn{1}{c|}{10} &
			Max Waypoint Dist. [m] &
			5 \\
			ICP Max Trans. Diff. [m] &
			\multicolumn{1}{c|}{0.01} &
			Loop Closure Radius [m] &
			\multicolumn{1}{c|}{3} &
			Resolution Radius [m] &
			\multicolumn{1}{c|}{1.5} &
			Max Velocity [\SI{}{\meter\per\second}] & 1  \\
			ICP Max Rot. Diff. [rad] & \multicolumn{1}{c|}{0.001} & Loop Closure Min Inliers & \multicolumn{1}{l|}{5000} & Target Density [points \SI{}{\per\meter\cubed}] & \multicolumn{1}{c|}{5} & Acceleration [\SI{}{\meter\per\second\squared}] &
			0.1  \\ \bottomrule
		\end{tabular}
	\end{adjustbox}
\end{table}

\begin{figure}[tp]
	\centering
	\captionsetup[subfigure]{}
	\captionsetup[subfigure]{labelformat=empty}
	\captionsetup[subfigure]{justification=centering}
	
	\subfloat[\colorbox{cyan!15}{\textbf{Industrial Building A (front)}}]{\includegraphics[width=.33\linewidth]{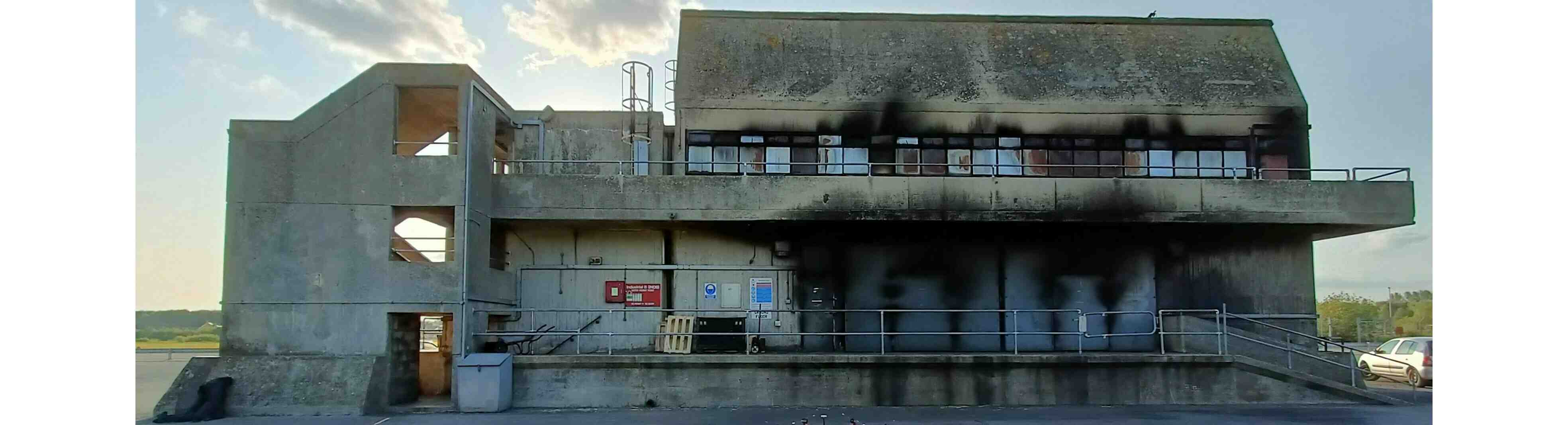}} \hfill
	\subfloat[\colorbox{magenta!15}{\textbf{Industrial Building A (back)}}]{\includegraphics[width=.33\linewidth]{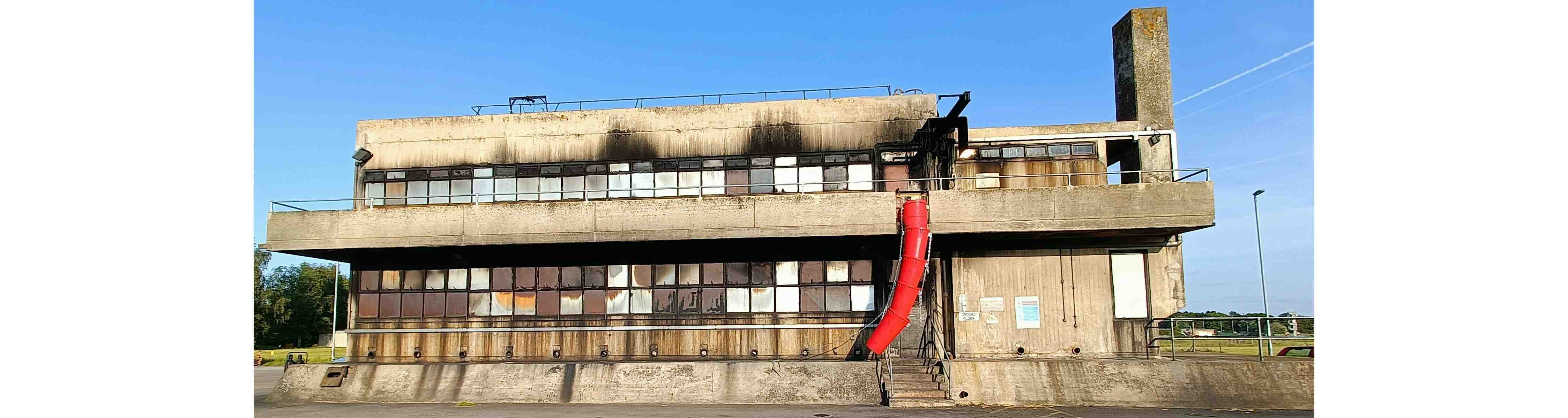}} \hfill
	\subfloat[\colorbox{yellow!15}{\textbf{Rig 5 (front)}}]{\includegraphics[width=.33\linewidth]{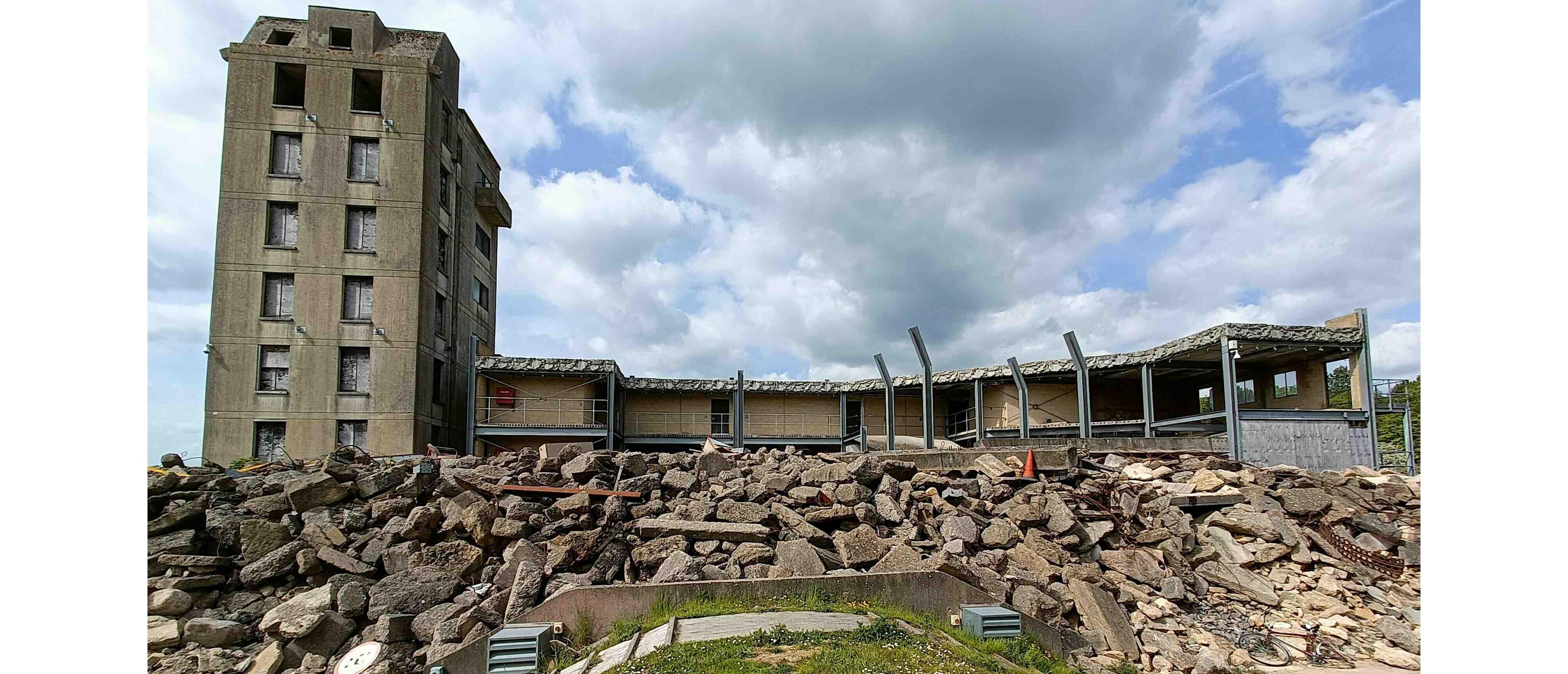}} 
	\vspace{-1ex}
	\subfloat[\colorbox{cyan!15}{Leica Map}]{\includegraphics[width=.33\linewidth]{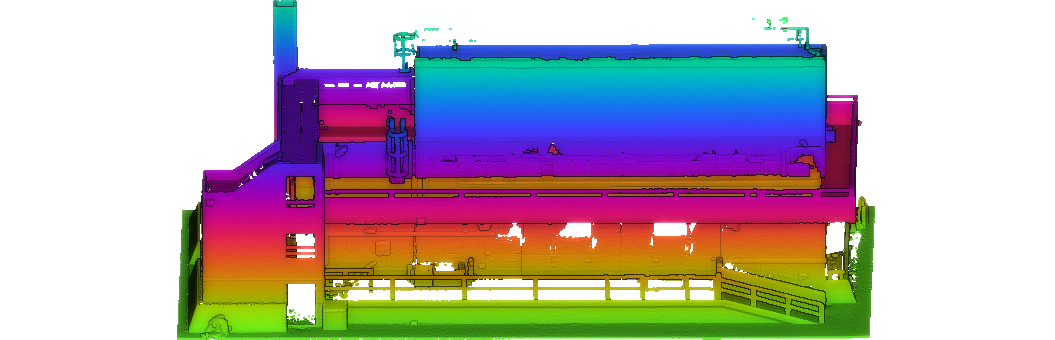}} \hfill
	\subfloat[\colorbox{magenta!15}{Leica Map}]{\includegraphics[width=.33\linewidth]{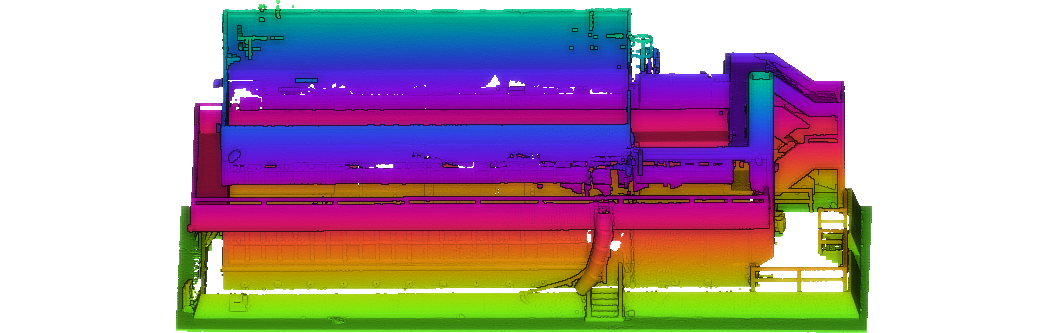}} \hfill
	\subfloat[\colorbox{yellow!15}{Leica Map}]{\includegraphics[width=.33\linewidth]{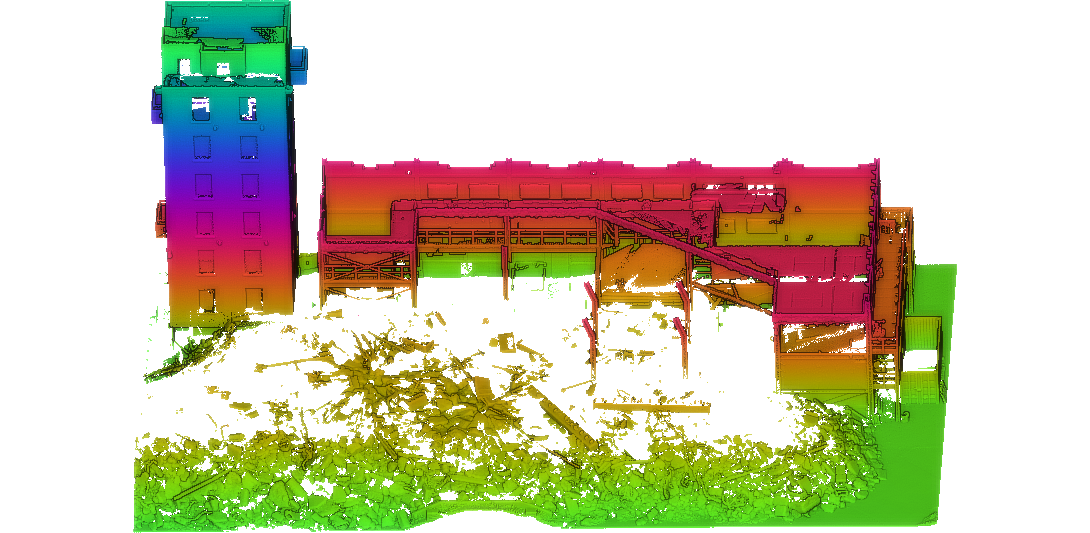}}
	\vspace{-1ex}
	\subfloat[\colorbox{cyan!15}{Pilot Map and Trajectory}]{\includegraphics[width=.33\linewidth]{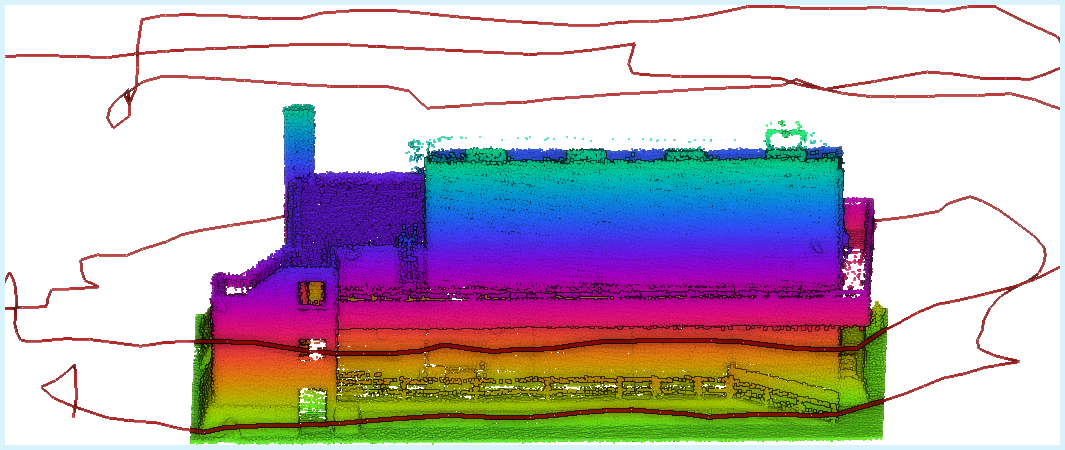}} \hfill
	\subfloat[\colorbox{magenta!15}{Pilot Map and Trajectory}]{\includegraphics[width=.33\linewidth]{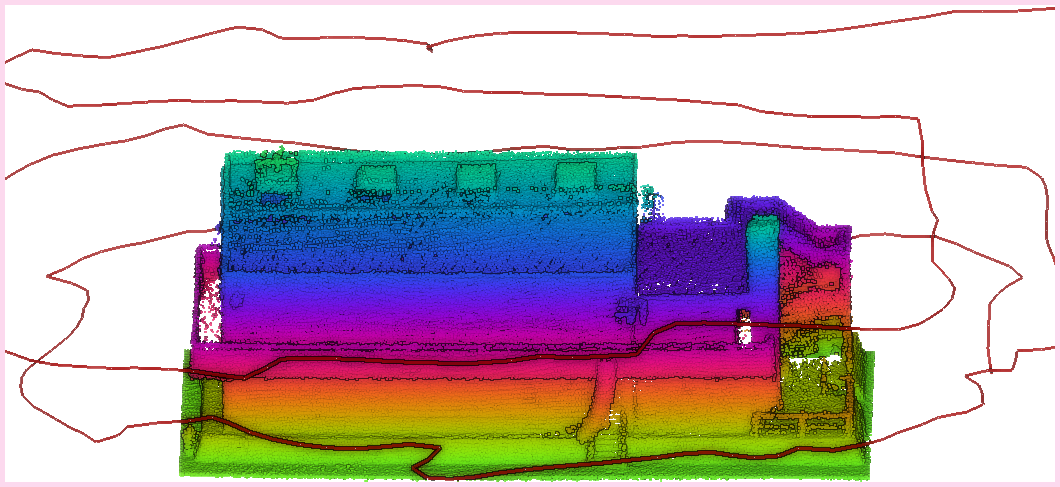}} \hfill
	\subfloat[\colorbox{yellow!15}{Pilot Map and Trajectory}]{\includegraphics[width=.33\linewidth]{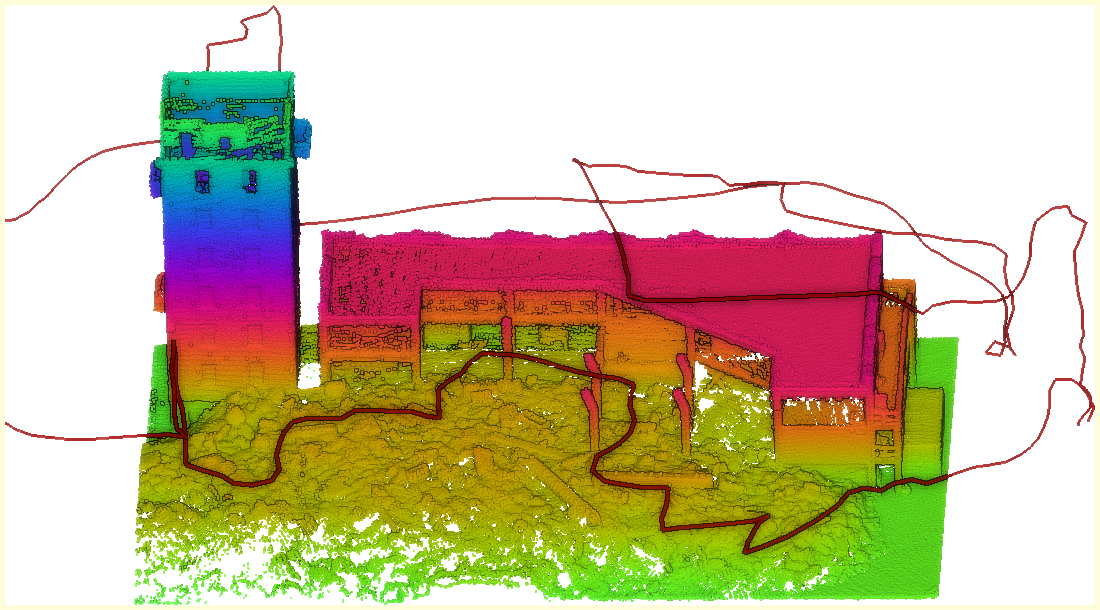}} 
	\vspace{-1ex}
	\subfloat[\colorbox{cyan!15}{Autonomy Map and Trajectories}]{\includegraphics[width=.33\linewidth]{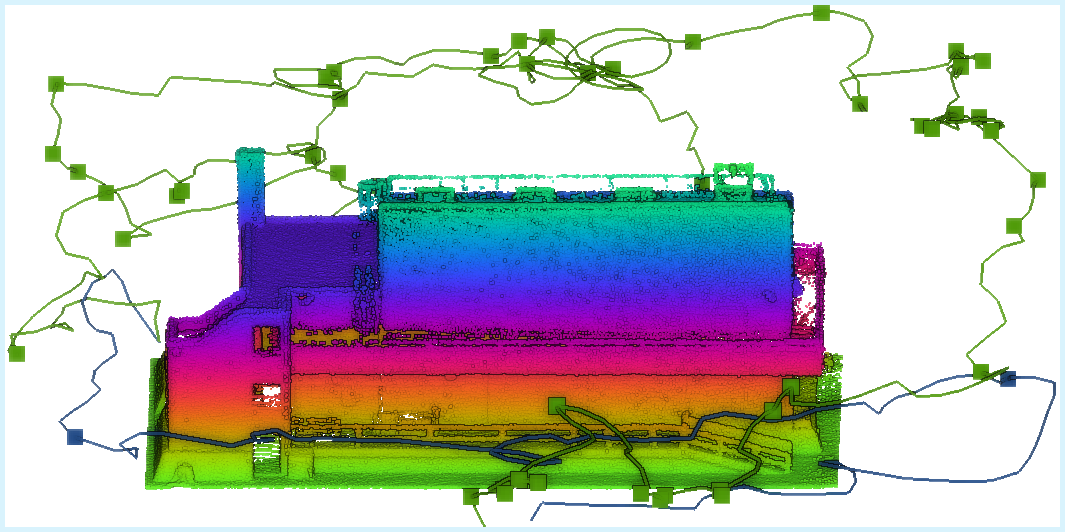}} \hfill
	\subfloat[\colorbox{magenta!15}{Autonomy Map and Trajectories}]{\includegraphics[width=.33\linewidth]{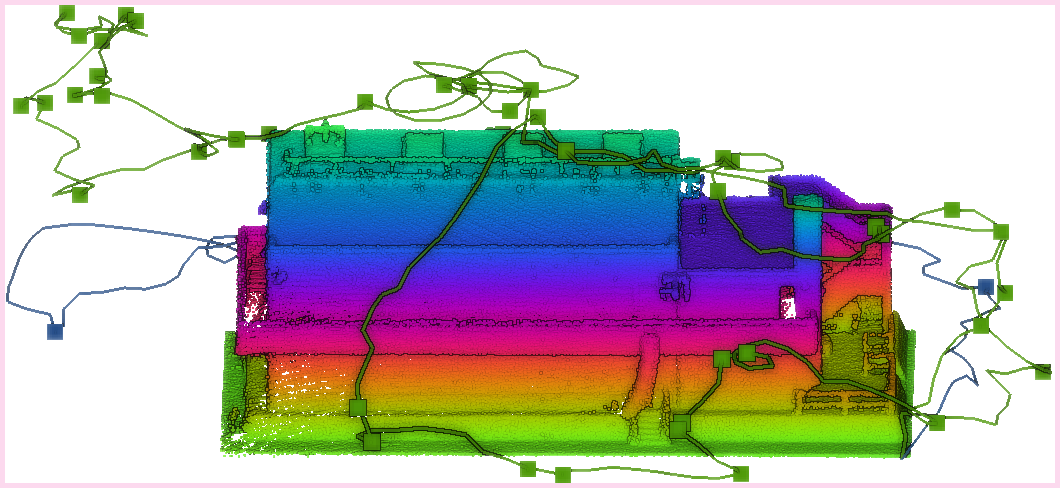}} \hfill
	\subfloat[\colorbox{yellow!15}{Autonomy Map and Trajectories}]{\includegraphics[width=.33\linewidth]{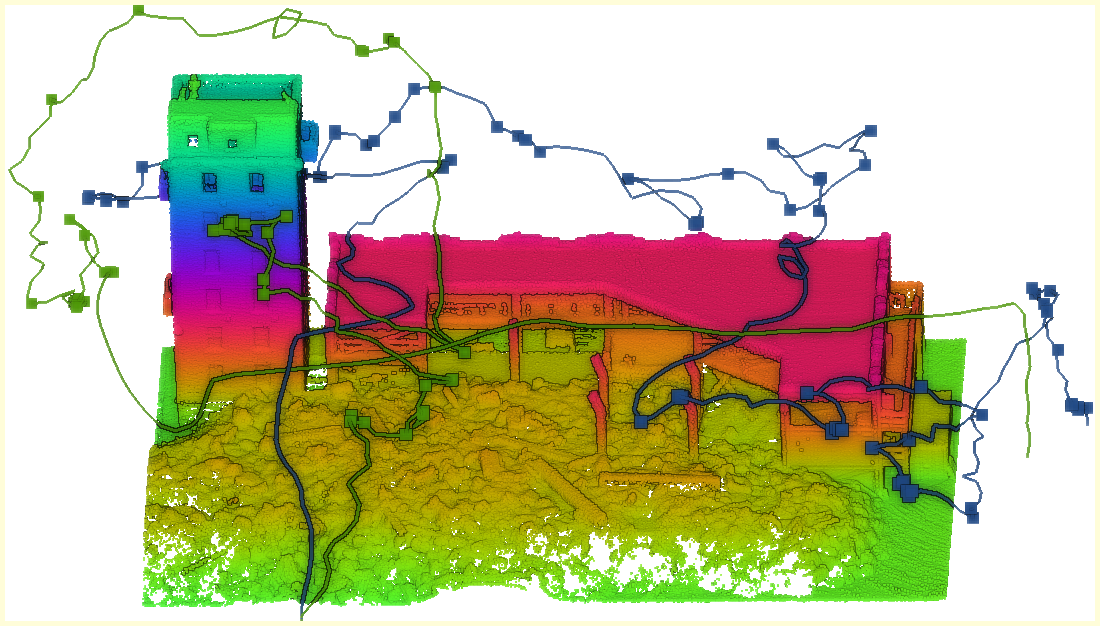}}
	\vspace{-1ex}
	\subfloat[\colorbox{cyan!15}{Pilot Map overlaid on Autonomy Map}]{\includegraphics[width=.33\linewidth]{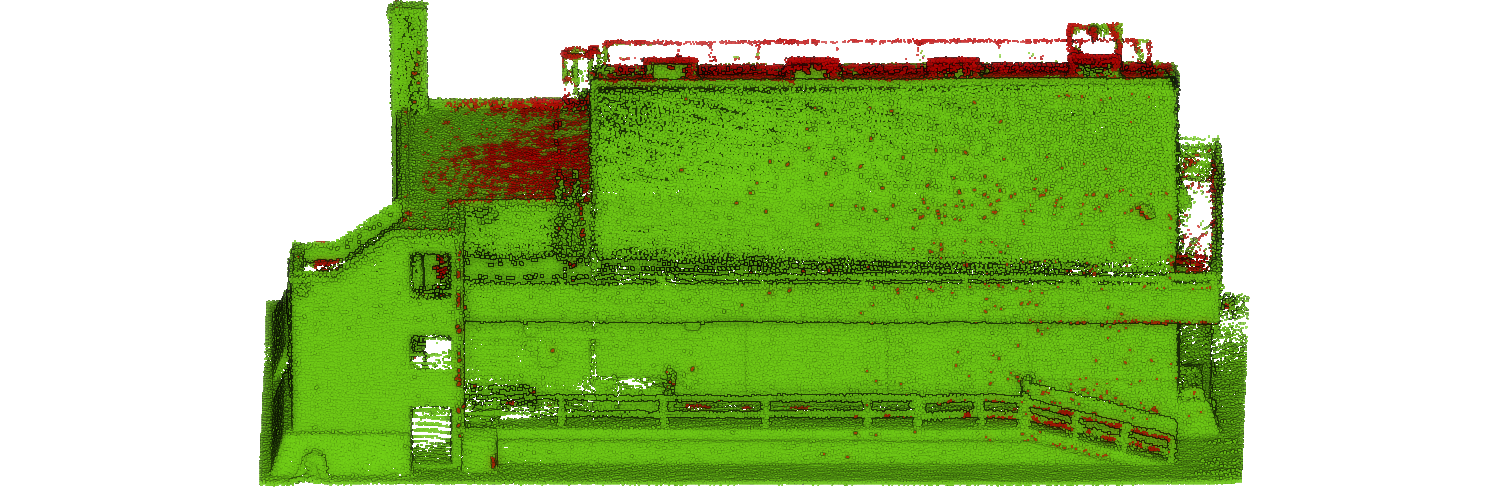}} \hfill
	\subfloat[\colorbox{magenta!15}{Pilot Map overlaid on Autonomy Map}]{\includegraphics[width=.33\linewidth]{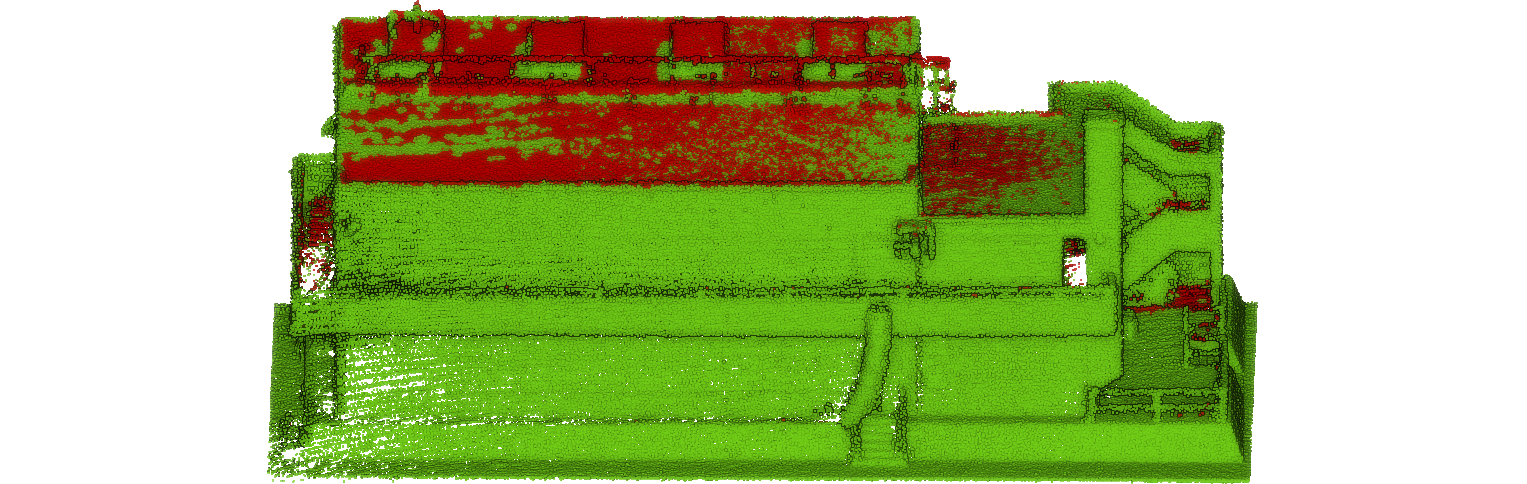}} \hfill
	\subfloat[\colorbox{yellow!15}{Pilot Map overlaid on Autonomy Map}]{\includegraphics[width=.33\linewidth]{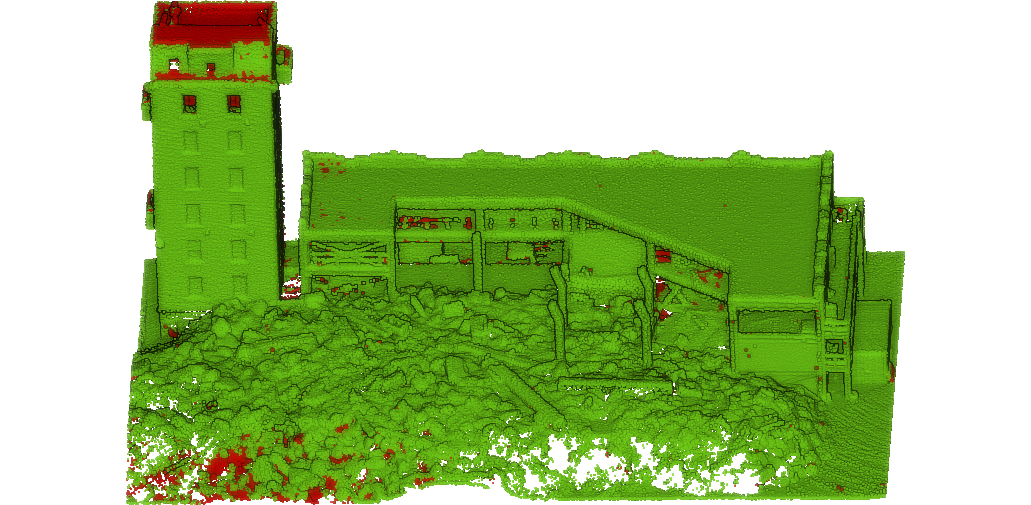}}
	\vspace{-1ex}
	\caption{Qualitative results showing the pointcloud maps obtained by field experiments conducted at Industrial Building A and Rig 5 (Autonomy 1 mission). The last row shows a comparison between the Autonomy map (green) and the Pilot map (red and green).}
	\figlabel{osprey-results-1}
\end{figure}

\begin{figure}[tp]
	\centering
	\captionsetup[subfigure]{}
	\captionsetup[subfigure]{labelformat=empty}
	\captionsetup[subfigure]{justification=centering}
	
	\subfloat[\colorbox{cyan!15}{\textbf{Rig 5 (back)}}]{\includegraphics[width=.33\linewidth]{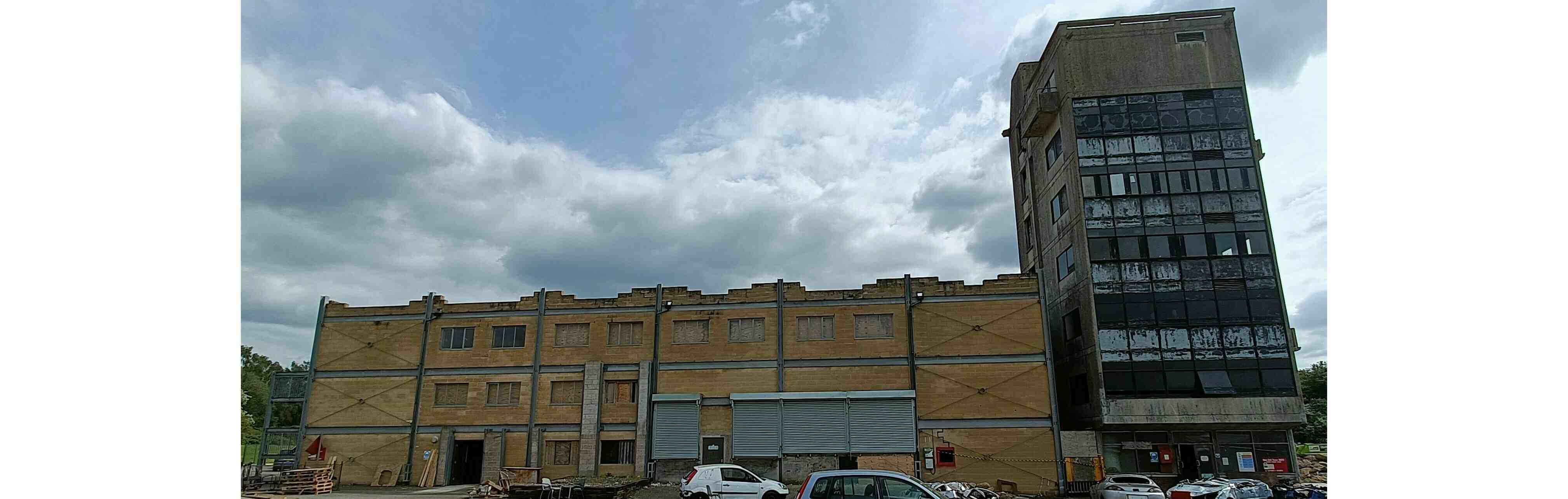}}	
	\subfloat[\colorbox{magenta!15}{\textbf{Industrial Building B (front)}}]{\includegraphics[width=.33\linewidth]{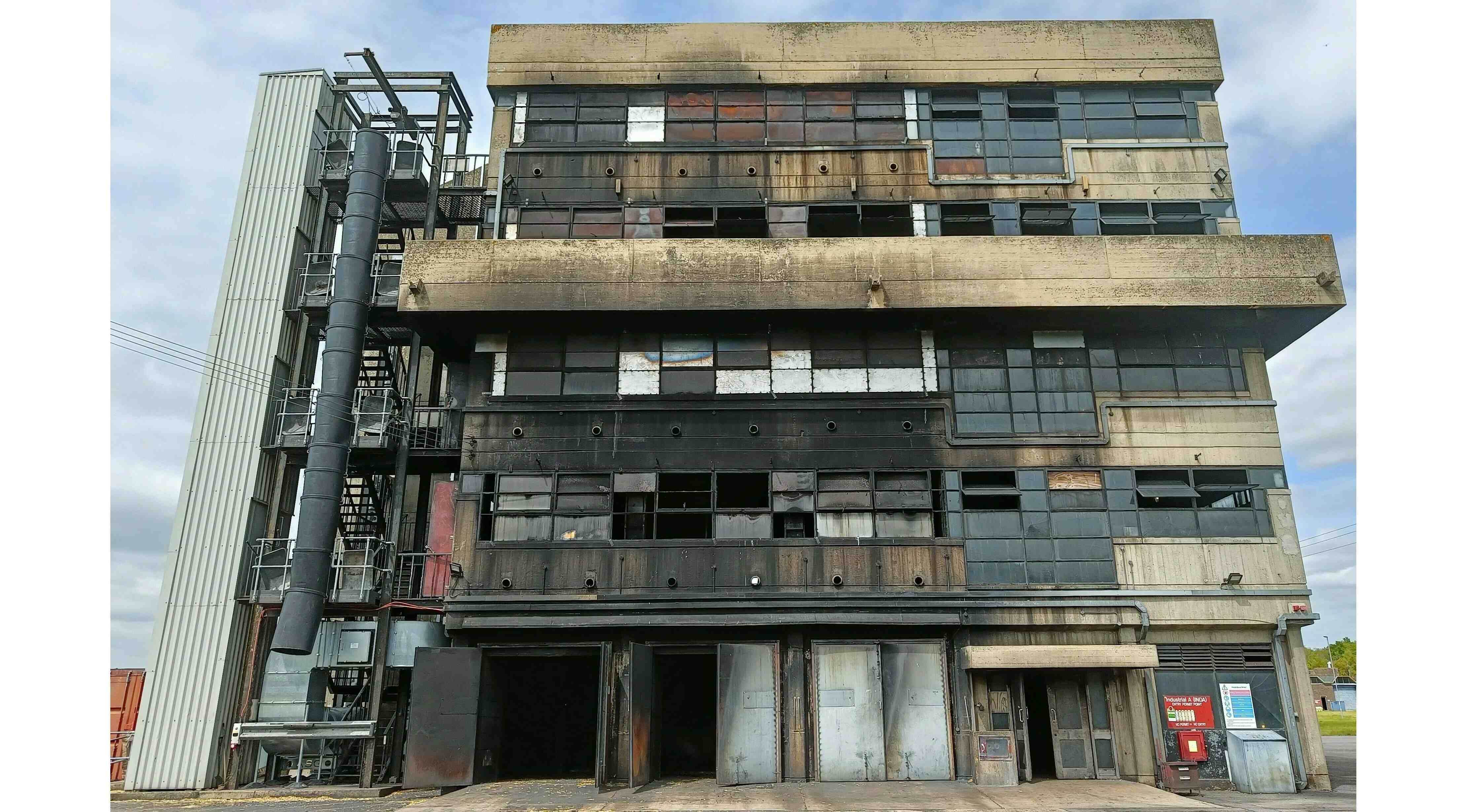}} \hfill
	\subfloat[\colorbox{yellow!15}{\textbf{Industrial Building B (back)}}]{\includegraphics[width=.33\linewidth]{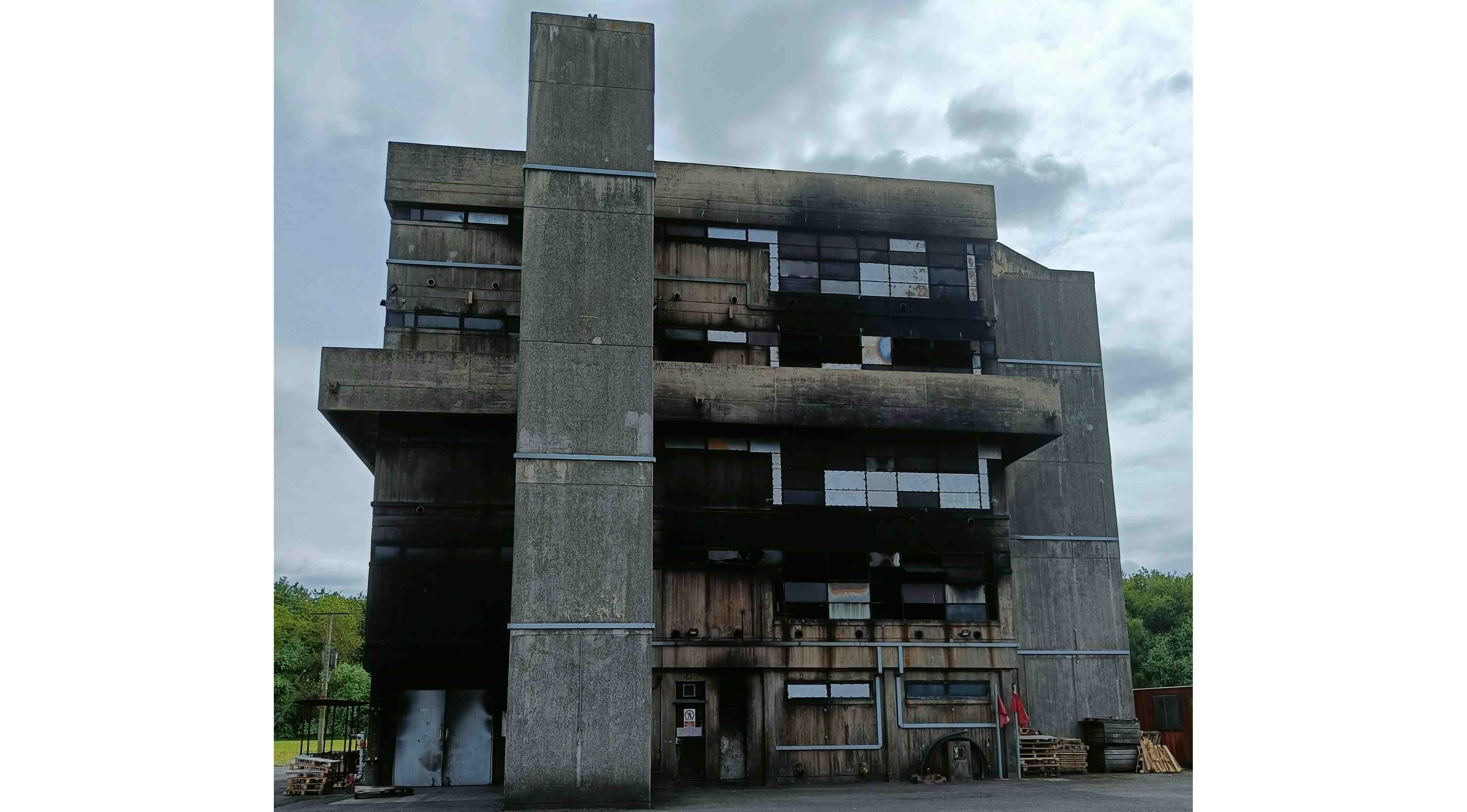}} \hfill
	\vspace{-1ex}
	\subfloat[\colorbox{cyan!15}{Leica Map}]{\includegraphics[width=.33\linewidth]{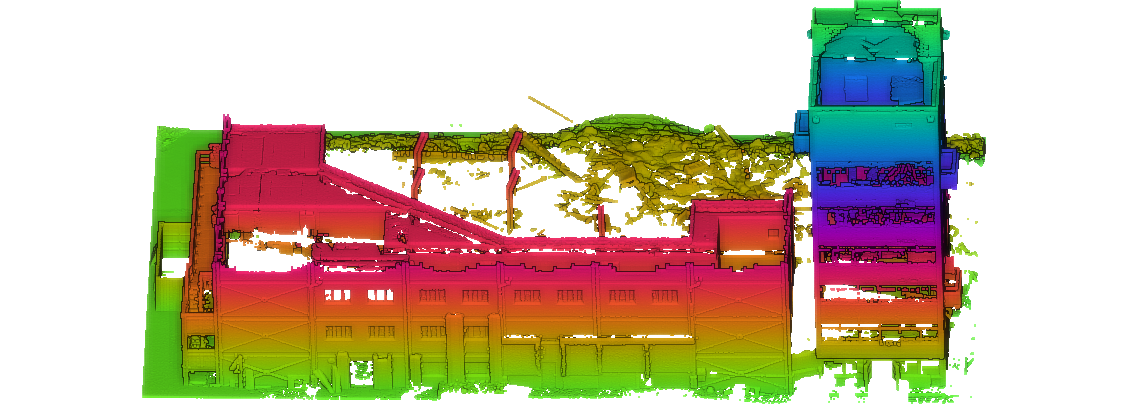}} 
	\subfloat[\colorbox{magenta!15}{Leica Map}]{\includegraphics[width=.33\linewidth]{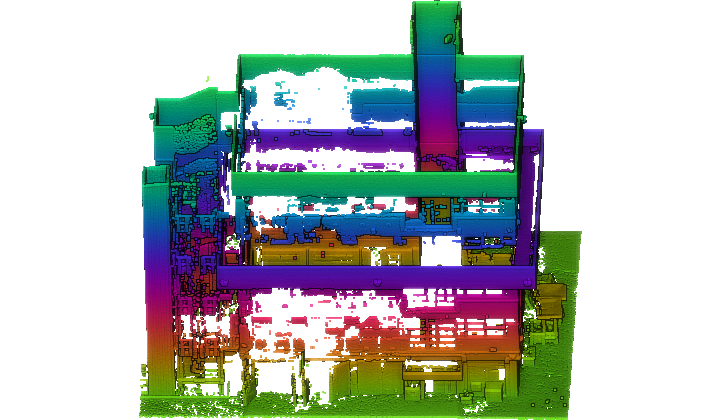}} \hfill
	\subfloat[\colorbox{yellow!15}{Leica Map}]{\includegraphics[width=.33\linewidth]{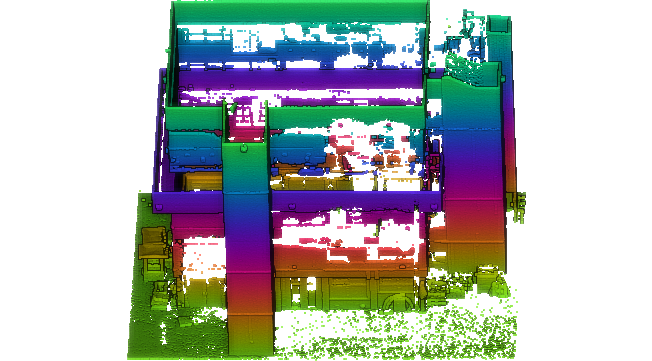}} \hfill
	\vspace{-1ex}
	\subfloat[\colorbox{cyan!15}{Pilot Map and Trajectory}]{\includegraphics[width=.33\linewidth]{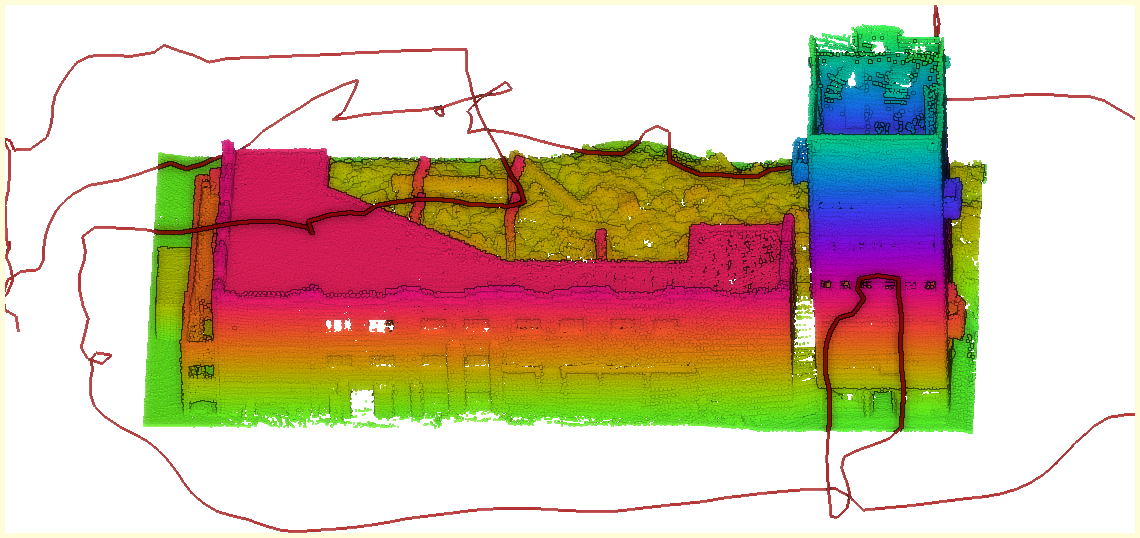}}
	\subfloat[\colorbox{magenta!15}{Pilot Map and Trajectory}]{\includegraphics[width=.33\linewidth]{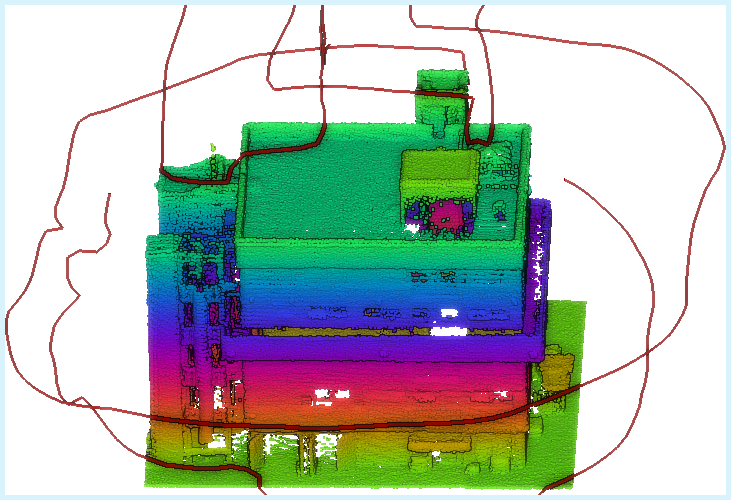}} \hfill
	\subfloat[\colorbox{yellow!15}{Pilot Map and Trajectory}]{\includegraphics[width=.33\linewidth]{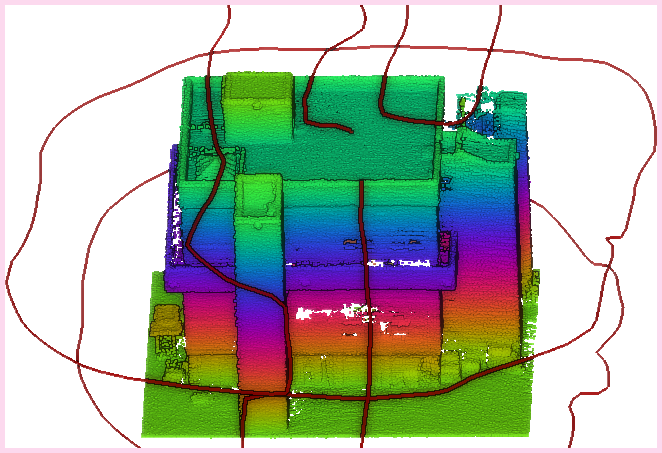}} \hfill
	\vspace{-1ex}
	\subfloat[\colorbox{cyan!15}{Autonomy Map and Trajectories}]{\includegraphics[width=.33\linewidth]{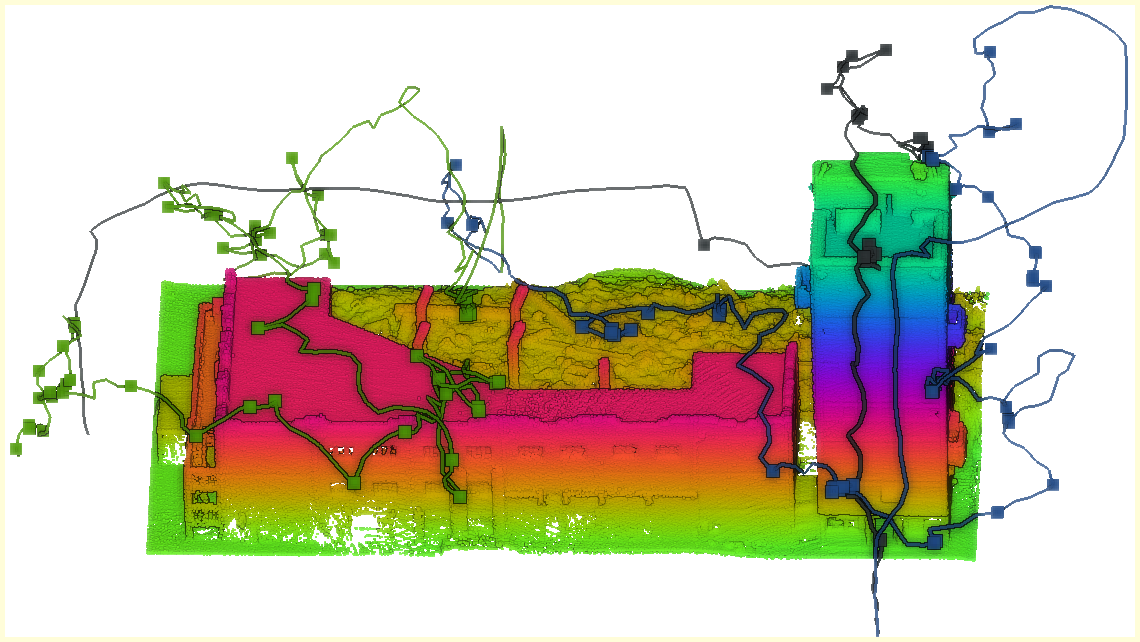}} 
	\subfloat[\colorbox{magenta!15}{Autonomy Map and Trajectories}]{\includegraphics[width=.33\linewidth]{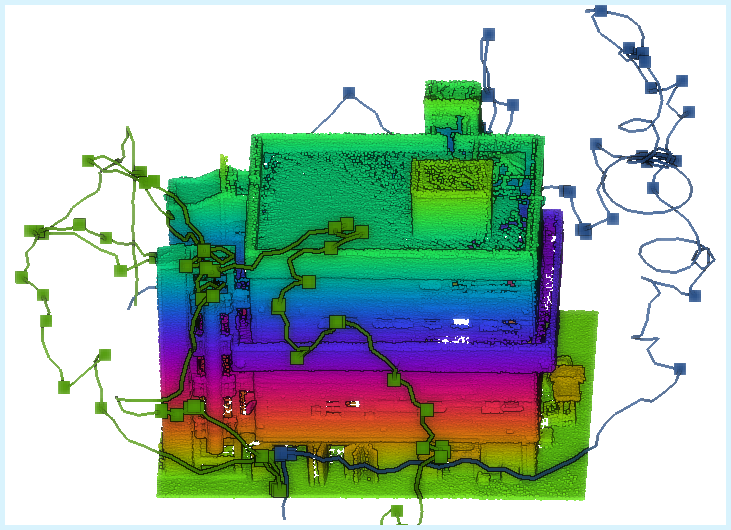}} \hfill
	\subfloat[\colorbox{yellow!15}{Autonomy Map and Trajectories}]{\includegraphics[width=.33\linewidth]{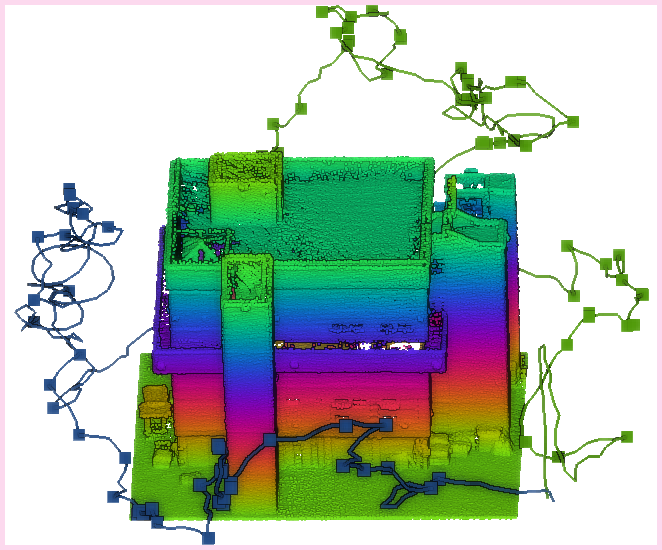}} \hfill
	\vspace{-1ex}
	\subfloat[\colorbox{cyan!15}{Pilot Map overlaid on Autonomy Map}]{\includegraphics[width=.33\linewidth]{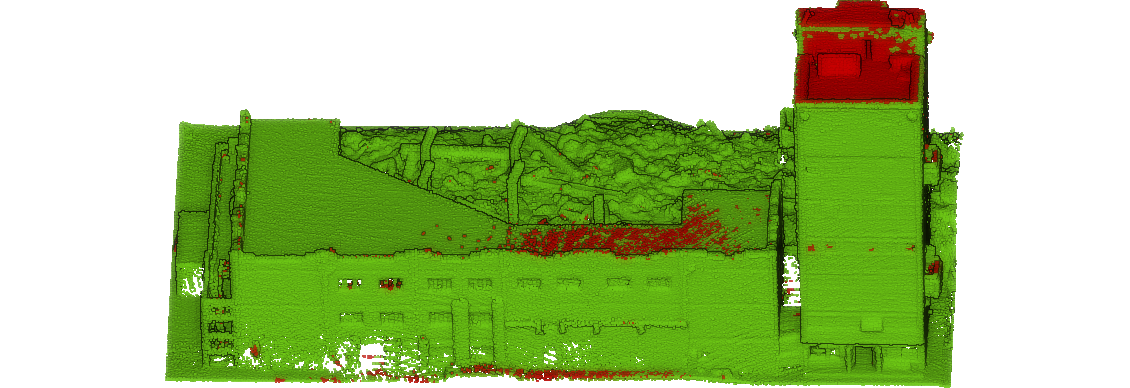}} 
	\subfloat[\colorbox{magenta!15}{Pilot Map overlaid on Autonomy Map}]{\includegraphics[width=.33\linewidth]{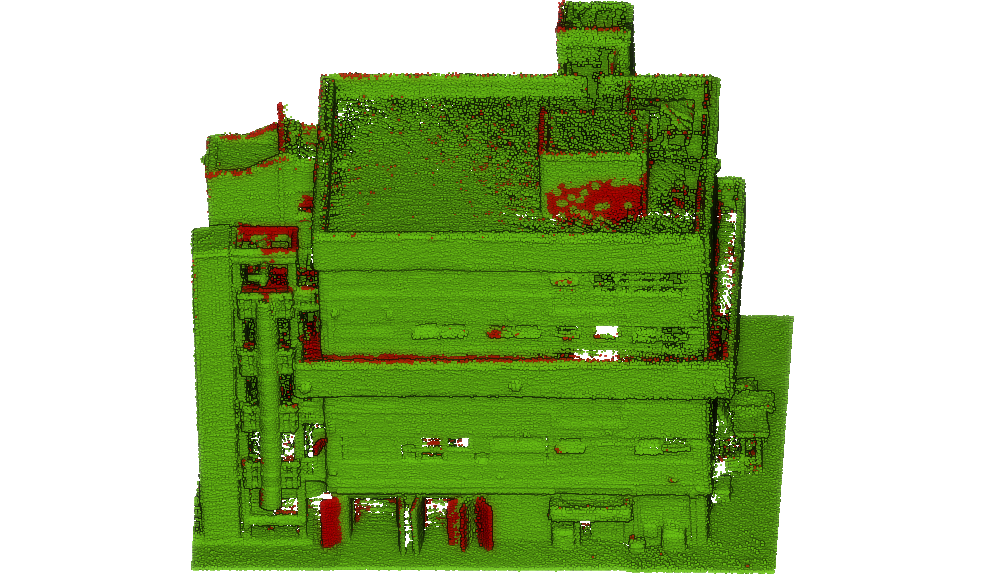}} \hfill
	\subfloat[\colorbox{yellow!15}{Pilot Map overlaid on Autonomy Map}]{\includegraphics[width=.33\linewidth]{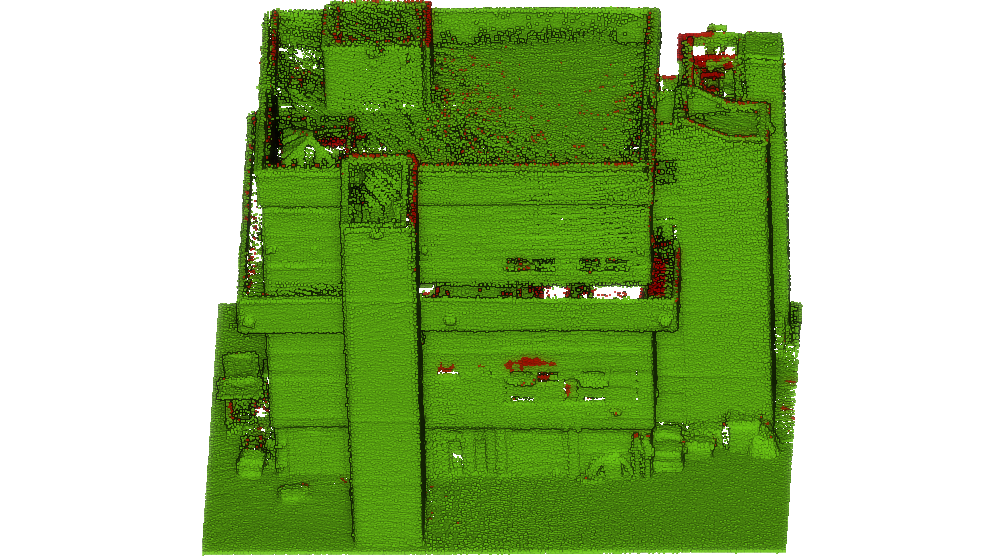}} \hfill
	\vspace{-1ex}
	\caption{Qualitative results showing the pointcloud maps obtained by field experiments conducted at Rig 5 (Autonomy 2 mission) and Industrial Building B. The last row shows a comparison between the Autonomy map (green) and the Pilot map (red and green).}
	\figlabel{osprey-results-2}
\end{figure}               

\section{Evaluation}
\seclabel{evaluation}

\renewcommand{\arraystretch}{1.1}
\begin{table}[]
	\centering
	\caption{Quantitative metrics for the field experiments, measuring the efficiency of each mapping approach and the quality of maps obtained at each site. Two autonomy missions were carried out at Rig 5.}
	\vspace{1ex}
	\tbllabel{results-table}
	\begin{adjustbox}{width=\linewidth,center}
		\aboverulesep=0ex
		\belowrulesep=0ex
		\begin{tabular}{@{}l|ccc|ccc|cccc@{}}
			\toprule
			\multicolumn{1}{c}{} & \multicolumn{3}{c}{Industrial Building A} & \multicolumn{3}{c}{Industrial Building B} & \multicolumn{4}{c}{Rig 5}               \\ \midrule 
			\multicolumn{1}{c|}{} & Leica       & Pilot       & Autonomy      & Leica       & Pilot       & Autonomy      & Leica & Pilot & Autonomy 1 & Autonomy 2 \\ \midrule
			Scans / Flights & 19 & \textbf{1} & 2 & 9 & \textbf{1} & 2 & 23 & \textbf{1} & 2 & 3 \\
			Travel Distance [m] & --- & \textbf{400} & 512 & --- & \textbf{476} & 541 & --- & \textbf{445}& 651 & 824 \\
			Mission Time [s]     & 7980 & \textbf{493} & 1956 & 3780 & \textbf{535} & 2271 & 9660 & \textbf{538} & 2498 & 3323 \\
			Flight Speed [\SI{}{\meter\per\second}] & --- & \textbf{0.81} & 0.26 & --- & \textbf{0.89} & 0.24 & --- & \textbf{0.83} & 0.26 & 0.25 \\
			View Planning Time [s] & --- & --- & 10 & --- & --- & 24 & --- & --- & 39 & 37 \\
			Path Planning Time [s] & --- & --- & 175 & --- & --- & 144 & --- & --- & 89 & 308 \\
			Coverage [\%] & 83.4 & 87.6 & \textbf{93.8} & 75.5 & 89.7 & \textbf{95.7} & 75.4 & 85.0 & 91.1 & \textbf{92.9} \\
			Accuracy [m] & --- & \textbf{0.04} & 0.05 & --- & 0.07 & \textbf{0.06} & --- & 0.08 & \textbf{0.06} & 0.09 \\
			\bottomrule
		\end{tabular}
	\end{adjustbox}
\end{table}

The mapping performance of Osprey is evaluated qualitatively and quantitatively based on the presented field experiments. We compare the quality and efficiency of its mapping with results obtained from the pilot-flown missions and Leica BLK360 \gls{tls}. 

Figures \ref{fig:osprey-results-1} and 12 present the pointcloud maps obtained by the mapping approaches for each of the three sites. Maps from each approach were aligned into a common reference frame using CloudCompare\footnote{\url{https://www.cloudcompare.org/}}. Each column presents the maps for one site, viewed from either the front or back. The first row presents a photograph of the site from the corresponding perspective. Rows 2--4 present the maps obtained by the Leica BLK360, pilot-flown mission and autonomous Osprey mission, respectively. These maps are coloured with a $z$-axis colourmap and also show the flight trajectories of the platform for the aerial missions. The squares along the autonomy trajectories denote the views captured by \gls{see}. The last row shows a coverage comparison between the aligned autonomy and pilot maps. This is computed by matching points in the autonomy map with their nearest neighbours in the pilot map. Green points have a match closer than $10$~cm and indicate good consistency between the maps. Red points have no match within $10$~cm and denote areas that were missed in the piloted mission.

\ttblref{results-table} presents quantitative metrics to evaluate the efficiency of each mapping approach and the quality of maps obtained. The travel distance measures the total distance flown by the aerial platform during a mapping mission, from takeoff to landing for each flight. The mission time is the total scanning time taken for each Leica BLK360 scan or the overall flight time for the aerial missions, from takeoff to landing for each flight. The flight speed is the average speed of the platform during the aerial missions. The view and path planning times are specific to the Osprey autonomy missions. They measure the total computation time taken by the mission and motion planning algorithms. The computation time of the odometry and mapping algorithms was not measured separately as they ran continuously in real-time for the duration of a mission.       

The quality metrics quantify the coverage and accuracy of the captured maps. Map coverage is measured by comparing the individual maps obtained by each approach with a combined map created by aggregating the pointclouds from all of the maps in a common reference frame. Points in the combined map are matched with their nearest neighbours in each individual map; those with a match closer than $10$~cm are considered covered and those without are uncovered. The coverage metric is computed as the percentage of points in the combined map that are covered by an individual map. The mapping accuracy of the pilot-flown and Osprey autonomy missions is evaluated relative to the Leica BLK360 maps as it provides survey-grade mapping accuracy. The accuracy metric measures the mean distance between points in the Leica BLK360 map of a site and their nearest neighbour matches in the pilot-flown or autonomy map.      

\subsection{Pointcloud and NeRF Colour Reconstructions}

\begin{figure}[tp]
	\centering
	\captionsetup[subfigure]{}
	\captionsetup[subfigure]{labelformat=empty}
	\captionsetup[subfigure]{justification=centering}
	
	\subfloat[Industrial Building A]{\includegraphics[width=.3\linewidth]{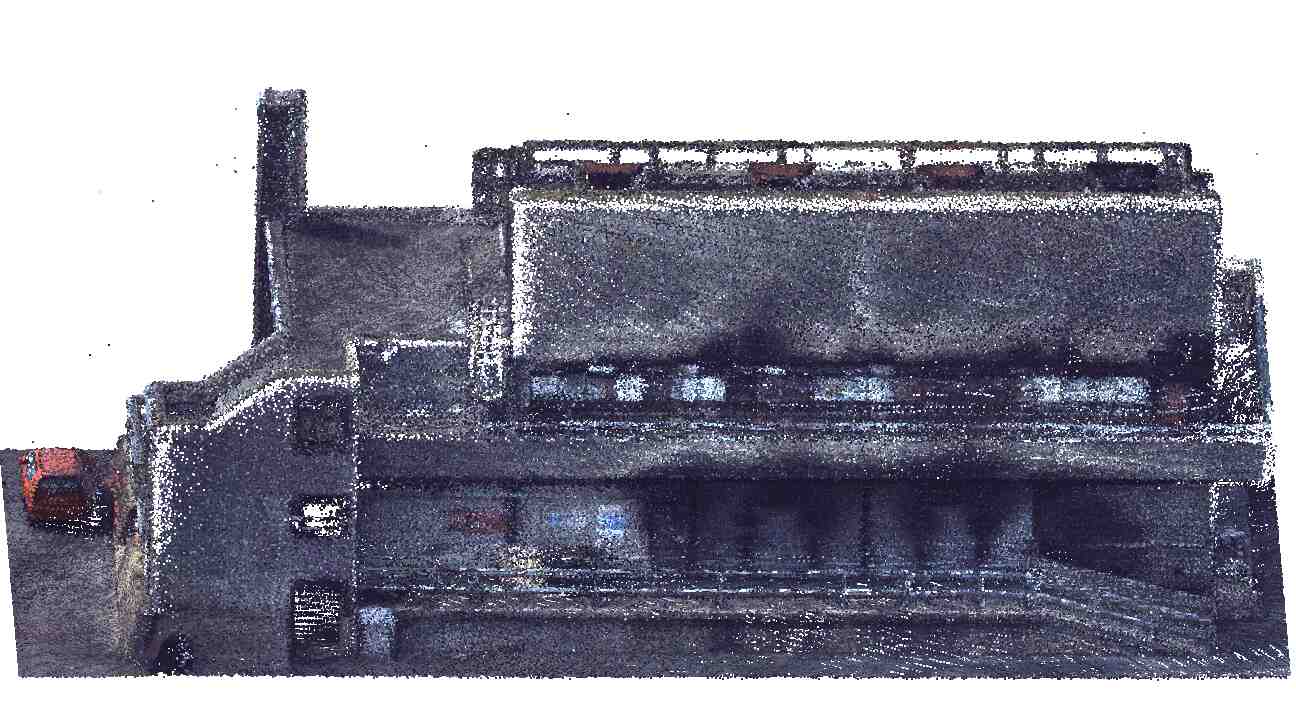}} \hfill
	\subfloat[Industrial Building B]{\includegraphics[width=.2\linewidth]{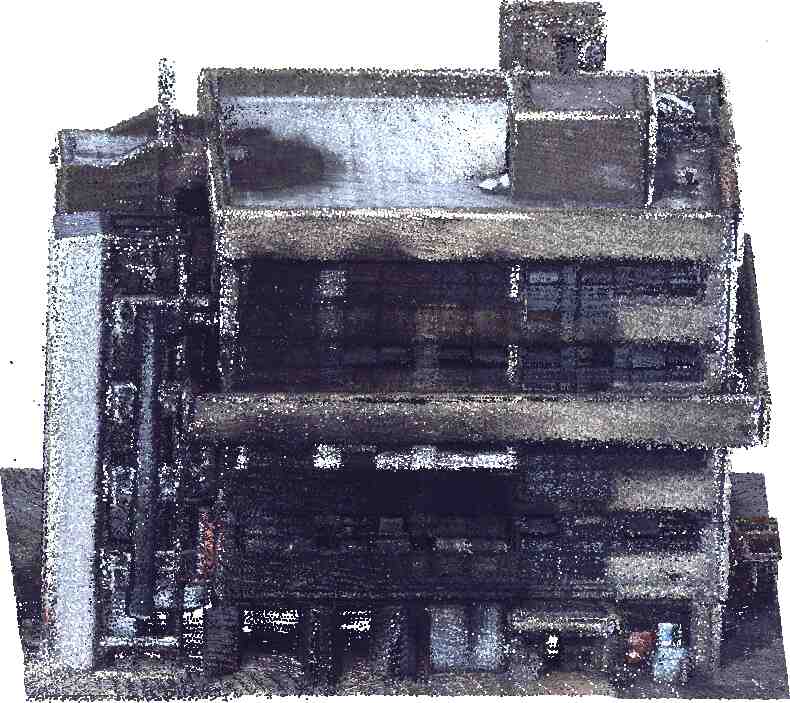}} \hfill
	\subfloat[Rig 5]{\includegraphics[width=.3\linewidth]{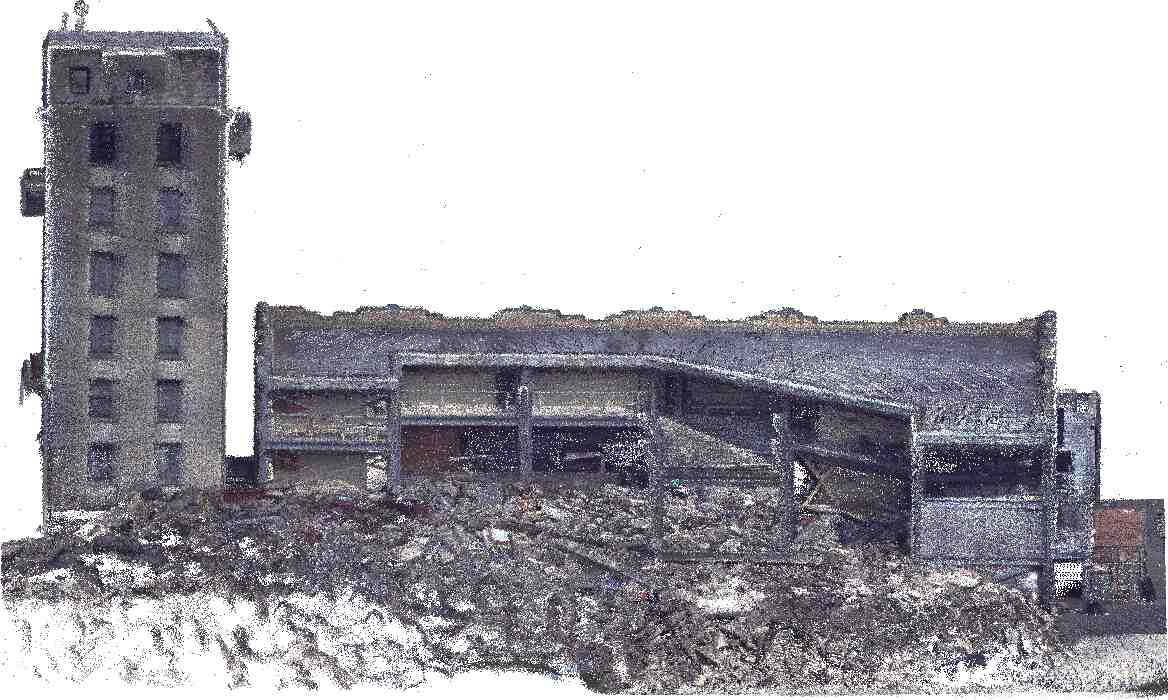}} 
	\vspace{-1ex}
	\caption{Pointcloud maps of each site, captured by the Osprey autonomy missions, that were colourised using the camera images. This colourisation was performed offline using a custom pipeline based on COLMAP.}
	\figlabel{osprey-colour-results}
\end{figure}

\begin{figure}[tp]
	\centering
	\includegraphics[width=0.9\linewidth]{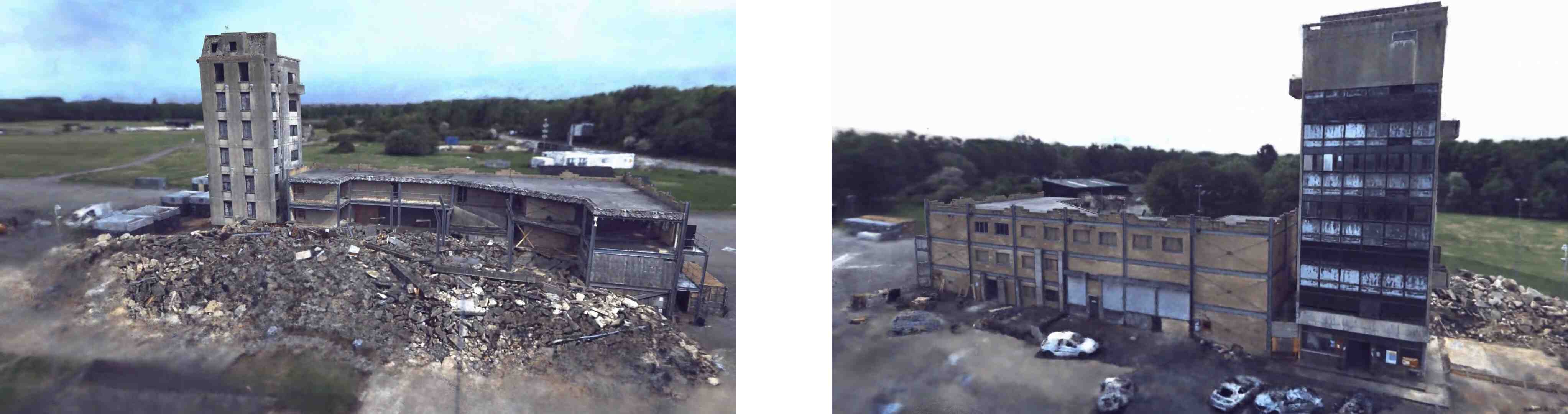}
	\vspace{-2ex}
	\caption{Two novel views of Rig 5 rendered by our \textit{SiLVR} NeRF-mapping system in post-processing.}
	\figlabel{rig-5-silvr}
\end{figure}

The LiDAR on the Frontier device overlaps with three fisheye colour cameras, which we used for offline true colour reconstruction. We considered two different methods: projecting colour from the camera images onto the pointcloud map \figref{osprey-colour-results} and producing a \gls{nerf} reconstruction using SiLVR\footnote{\url{https://ori-drs.github.io/projects/silvr/}} \citep[\tfigref{rig-5-silvr};][]{Tao2023}.

The projected colour reconstructions were created with a custom pipeline, which uses COLMAP\footnote{\url{https://colmap.github.io/}} \citep{Schonberger2016} to optimise the SLAM-estimated camera poses and intrinsic parameters. The camera images are then projected onto the pointcloud map from their optimised poses to produce a true colour map that preserves the metric accuracy of the original map. When colour is projected onto a point from multiple images the result is averaged and points without projected colour are removed. This does not produce photorealistic results due to variable image exposure (e.g., along the building edges in \tfigref{osprey-colour-results}).

SiLVR creates \gls{nerf} reconstructions by using the raw LiDAR pointclouds, camera images and pose graph trajectory from VILENS-SLAM. The \gls{nerf} technique for differentiable volume rendering was first introduced by \citet{Mildenhall2020}. Our approach builds upon Nerfacto \citep{Tancik2023} by adding LiDAR measurements to regularise the depth and surface normals in the \gls{nerf} using LiDAR range images. In particular this improves the planarity of uniformly textured surfaces. The poses from VILENS-SLAM are used to initialise COLMAP, which refines them in a bundle adjustment. Large sites can be reconstructed by segmentation into submaps using Spectral Clustering \citep{VonLuxburg2007}. SiLVR produces photorealistic reconstructions with high fidelity and can achieve promising metric accuracy due to the LiDAR regularisation.        

\section{Discussion}
\seclabel{discussion}

The results (Figs. \ref{fig:osprey-results-1} and \ref{fig:osprey-results-2}, Table \ref{tbl:results-table}) show that the Osprey autonomous mapping system obtained the greatest map coverage for all the sites. This clearly demonstrates the value of mapping sites with an aerial platform and using an autonomous system with online mission planning capabilities. The Leica BLK360 is clearly unable to map areas that could not be seen from the ground such as the roofs of the buildings. We did not have access to a Leica BLK2FLY drone, which may address these limitations.

Our key demonstration is that the Osprey autonomy missions obtained greater coverage than the pilot-flown missions. This difference is most significant on the roofs and balconies of the buildings (Figs. \ref{fig:osprey-results-1} and \ref{fig:osprey-results-2}, last row). The pilot had to fly the platform high over the building roofs in order to maintain the direct line-of-sight visibility needed for safe operation and had no knowledge of structures on the roofs that were not visible from the ground. This made it challenging to determine if they had been completely mapped. The balcony interiors were difficult to map, particularly for the pilot, as from most angles they were occluded by the balcony walls and it was challenging for the pilot to determine from their perspective on the ground if the interiors were visible to the LiDAR. The results show that many of these limitations for the pilot-flown missions can be overcome by using a state-of-the-art autonomous aerial mapping system such as Osprey.

In terms of mapping efficiency, Osprey took less time than the Leica scanner surveys but took longer and flew further than the pilot-flown missions \ptblref{results-table}. The Leica scanner captured measurements at several locations around a site with each scan taking $7$ minutes. The Leica surveys took an average of three times longer than the Osprey autonomy missions. The difference in mission times between the Osprey autonomy and pilot-flown missions can be attributed to a number of factors. The Osprey autonomy system held position while performing view and path planning but this computation time was relatively insignificant --- only $1$\% of the overall mission time for view planning and $8$\% for path planning. A more significant factor is the difference in average flight speed, which was three times slower than for the pilot-flown missions. Osprey was constrained to fly slower than the piloted missions to mitigate the risk of the platform deviating from its planned trajectory when flying near large structures, where limited GPS connectivity could degrade the accuracy of the DJI flight controller state. This slower flight speed provided the safety pilot with enough time to safely prevent collisions.

The difference in travel distance and trajectory smoothness between the Osprey autonomy and pilot-flown missions can be attributed to the use of \emph{a priori} site knowledge for the piloted missions. The pilot preplanned their flights by walking around a site before takeoff and empirically judging what trajectory would provide the best coverage. This preplanning allowed the pilot to fly smoother and shorter trajectories that could be completed in a single flight. Incrementally planning potential views online with \gls{see} enabled Osprey to obtain greater map coverage than the pilot-flown missions, while achieving a similar map accuracy, but produced longer and less smooth trajectories as the mission planning algorithm made locally optimal decisions based on incomplete site knowledge. This could potentially be improved by planning a sequence of next best views instead of a single view and applying a smoothness constraint on the trajectory between them.   

The travelling required to land and takeoff between flights also increased the overall distance of the multi-session Osprey missions. We include the distance required to manually fly the platform to a safe landing location after reaching a critical battery level and the distance autonomously flown by the platform to resume its mission after replacing the batteries. This distance was sometimes significant, particularly for larger Rig 5 site, as when the drone was flying high above the opposite side of the building from the safety pilot they would have to fly the platform over the building while maintaining sufficient height for good visibility and identify a nearby location clear of obstacles for landing. It accounted for $23$\% of the overall travel distance in the second autonomy mission at Rig 5, which required two battery changes. This distance could be reduced in future work by autonomously identifying a safe landing location \citep[e.g.,][]{Bartolomei2022}. 

The overall performance of Osprey demonstrates the successful integration of its constituent algorithms. The \gls{vilens} odometry algorithm provided reliable pose estimates that facilitated the accurate registration of LiDAR measurements into a graph-based SLAM map by VILENS-SLAM, which uses loop closures to correct for odometry drift. The \gls{see} mission planning algorithm directed the platform to improve map coverage by capturing measurements from chosen views and collision-free paths between these views were planned by the \gls{aitstar} motion planning algorithm. LiDAR-based place recognition with ScanContext made the multi-session mapping possible by relocalising the platform within the existing SLAM map for an ongoing mission. This was the only combination of approaches tried when developing Osprey as they were well suited to the task of mapping large outdoor structures. All of the algorithms were able to run concurrently on the Frontier device without bottlenecks. The average computational load for \gls{vilens} was $\sim28\%$ of the CPU and for VILEN-SLAM this was $\sim1.8\%$. \gls{see} typically used $\sim 25\%$ of the CPU when processing new measurements and \gls{aitstar} used $\sim3.1\%$ when planning paths. The efficiency of the system could be further improved by sharing a single map representation between the algorithms. The performance of the individual algorithms is discussed in detail in the following subsections.

\subsection{Sensor Payload}

The perception capabilities of the aerial platform were well suited for mapping the target sites. The wide vertical field-of-view LiDAR made it possible to fly above tall structures and acquire more complete maps. The main limitation of this LiDAR is its maximum range of $20$~m, which made it necessary to fly relatively close to structures for robust odometry and could make it challenging to map sites with large gaps between structures. Newer LIDAR models are likely to relax this trade-off in the future.

An alternative design could have been to use multiple LiDARs mounted at different angles, a hemispherical LiDAR or a spinning LiDAR configuration \citep[e.g.,][]{Hudson2022} to achieve more complete coverage. Cameras with a greater dynamic range would provide better image exposure for visual odometry and pointcloud colouring --- in particular to avoid overexposure of surface edges facing the sky \figref{osprey-colour-results}.  

\subsection{Odometry}

The \gls {vilens} odometry algorithm provided a robust incremental motion estimate during the field experiments by tightly integrating high-frequency \gls{imu} measurements and lower frequency LiDAR measurements in a factor graph. The wide vertical field-of-view of the LiDAR ensured that there was always sufficient coverage of unique geometric structures for the \gls{icp} registration module to reliably converge to a good solution. All of the mapped sites were standalone structures, even though some contained multiple connected buildings, and therefore there was always enough 3D structure within range of the LiDAR to provide robust odometry. This would not necessarily be the case when mapping larger sites that contain disconnected structures. In this scenario it may be beneficial to include visual feature tracking \citep{Wisth2021b} or GPS measurements \citep{Beuchert2023} in the \gls{vilens} odometry estimate, both of which have been successfully demonstrated on other platforms in different environments.

\subsection{Mapping}

\begin{figure}[tp]
	\centering
	\includegraphics[width=.9\linewidth]{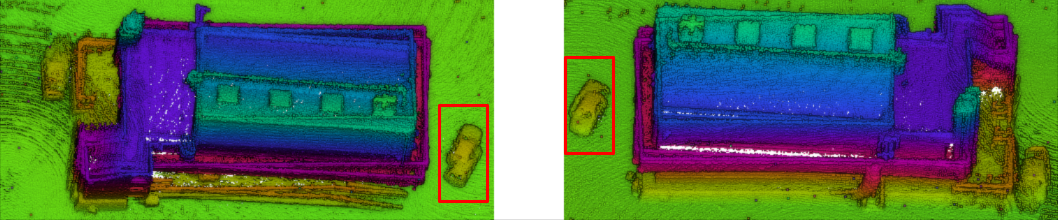}
	\vspace{-2ex}
	\caption{Two views of a pointcloud map with SLAM loop closure correction \emph{disabled} at Industrial Building A. It illustrates that pointcloud inconsistency occurs when there is uncorrected odometry drift (e.g., the misaligned car shown in the red boxes) and highlights the importance of highly accurate online SLAM for autonomous mapping.}
	\figlabel{loop-closure-failure}
\end{figure}

The VILENS-SLAM algorithm produced high quality pointcloud maps of the sites by incorporating geometric loop closures to correct for drift in the VILENS odometry estimate. The robustness of these loop closures was aided by low drift in the VILENS odometry, which in turn required the place recognition to find less frequent and more confident loop closure proposals. The importance of loop closure corrections is demonstrated in \tfigref{loop-closure-failure}, where the loop closure mechanism was disabled and this produced misalignments in the pointcloud map (e.g., the misaligned car shown in the red boxes).  

For multi-session mapping \psecref{multimap}, the ScanContext algorithm was usually reliable at relocalising the platform within a prior map during takeoff. Occasionally it was necessary for the safety pilot to fly the platform around the area of the existing map until a place recognition match could be identified. This occurred during test flights at Industrial Buildings A and B, where the sides of the building were sometimes not distinctive enough for a valid match and it was necessary to fly to a different location to find one. Relocalisation robustness could also be improved with visual place recognition \citep[e.g., ][]{Maffra2018}.       

\subsection{Mission Planning}

The \gls{see} mission planner reliably directed the platform to capture highly complete maps of the target sites. It was able to capture measurements from regions not covered by the pilot-flown missions or Leica surveys (e.g., the building roofs). It generally obtained better coverage of the balconies and other surfaces with restricted visibility but sometimes failed to capture measurements from areas that were considered unobservable after too many unsuccessful attempts. Some of these areas could possibly have been successfully observed at a closer range. Therefore, adding the ability to adapt the view distance after a chosen view fails to capture new measurements around a target frontier point may improve the coverage attainable by \gls{see}.

\gls{see} plans potential views to incrementally expand the coverage of measurements on visible surfaces. This works well when the surfaces in a target area are interconnected and coverage can be extended in all directions until a complete map is obtained. However, larger sites often contain multiple structures that are only connected by the ground. \gls{see} could observe these sites by including the ground in its target area but this may not be the most efficient method for extending coverage between structures. Future work could investigate adding a volumetric exploration strategy to \gls{see} for planning missions at sites with unconnected structures, which may be more efficient than planning solely based on measurement coverage. Adding an adaptive view distance would also be beneficial for observing surfaces in narrow gaps between structures.

\subsection{Motion Planning and Control}
\seclabel{motion-disc}

The \gls{aitstar} path planning algorithm could reliably plan efficient trajectories for the platform to traverse between chosen views. However, the executed trajectory sometimes deviated from the planned trajectory by as much as $1$~m due to poor performance of the DJI flight controller, which separately estimates its own state using onboard \gls{imu} and GPS sensors. The accuracy of this flight controller state is dependent on the number of visible GPS satellites and can degrade significantly when the platform flies close to large structures.  

Our control interface generated velocity commands for the DJI flight controller using the VILENS odometry, as this was more reliable, but the DJI flight controller sometimes produced different velocities than were commanded when its own internal state was inaccurate (e.g., due to low GPS accuracy). This reduced the safe flight speed of the drone and we needed to set maximum velocity and acceleration limits that were lower than desired. Switching to a flight controller that accepts input from an external state estimator (i.e., VILENS), such as PX4\footnote{\url{https://px4.io/}}, would enable closed-loop control based on a single reliable state estimate. 

\section{Conclusion}
\seclabel{conclusion}

This paper presents Osprey, an autonomous aerial mapping system with state-of-the-art multi-session mapping capabilities. Osprey integrates existing state-of-the-art approaches for multi-sensor odometry, LiDAR-based SLAM, mission planning, motion planning, and multi-session relocalisation into a complete pipeline capable of autonomously mapping large structures over multiple flights. The key contributions of this work are the development of Osprey, a demonstration of its mapping capabilities on several large outdoor sites and a discussion of outstanding challenges that should be addressed when developing future systems. A dataset\footnote{\url{https://dynamic.robots.ox.ac.uk/datasets/oxford-osprey}} of the field experiments has been made publicly available to benefit future research.       

The field experiments conducted with Osprey demonstrate that it is able to autonomously map large industrial sites and achieves greater coverage than pilot-flown missions and surveys with a \gls{tls}. It is quicker than carrying out a \gls{tls} survey and has equivalent map accuracy to the pilot-flown missions. Future improvements to the system could also allow Osprey to autonomously map sites at similar flight speeds to a human pilot and with comparable travel distances. Another promising direction for future work, beyond improving the individual Osprey system, would be to extend these capabilities to multiple platforms, which would enable the collaborative mapping of even larger and more complex sites.    

\subsubsection*{Acknowledgments}
The authors would like to thank Wayne Tubby, Matt Towlson and Arundathi Shanthini at ORI for building the Frontier sensor payload and mounting infrastructure for the DJI M600. This research was funded by Horizon Europe through the Digiforest project [101070405] and by UK Research and Innovation and EPSRC through ACE-OPS: From Autonomy to Cognitive assistance in Emergency OPerationS [EP/S030832/1].

\bibliographystyle{apalike}
\bibliography{library}

\end{document}